\documentclass[10pt,twocolumn,letterpaper]{article}

\usepackage[pagenumbers]{cvpr} 

\definecolor{cvprblue}{rgb}{0.21,0.49,0.74}
\usepackage[pagebackref,breaklinks,colorlinks,allcolors=cvprblue]{hyperref}

\usepackage{booktabs}
\usepackage{amsfonts}
\usepackage{multirow}
\usepackage{graphicx}
\usepackage{amsmath}
\usepackage{nicefrac}
\usepackage{microtype}
\usepackage{caption}
\usepackage{float}
\usepackage{placeins}
\usepackage{pifont}
\newcommand{\cmark}{\ding{51}}
\newcommand{\xmark}{\ding{55}}

\def\confName{CVPR}
\def\confYear{2027}

\title{SplatFlux: Feed-Forward 4D Gaussian Reconstruction and Generation \\ from Shared Video Latents}

\author{
	Runmin Zhang$^{1, 2}$\footnotemark[1]\hspace{4pt}\footnotemark[2]\quad
    Binglin Ren$^{1, 2}$\footnotemark[1]\hspace{4pt}\footnotemark[2]\quad
    Zhu Yu$^{1}$\footnotemark[1]\quad
    Fudong Wang$^{2}$\quad
    Jiangke Lin$^{2}$\\
    Yichun Shentu$^{2}$\quad
    Si-Yuan Cao$^{1}$\quad
    Yi Yuan$^{2}$\footnotemark[3]\quad
    Hui-Liang Shen$^{1}$\footnotemark[3]\\
	{\small $^{1}$Zhejiang University}\quad
	{\small $^{2}$Ant Group}\\
    {\tt\small \href{https://rm-zhang.github.io/SplatFlux/}{Project Page: https://rm-zhang.github.io/SplatFlux/}}
}

\DeclareMathSizes{8.75}{8.75}{6}{5}

\begin{document}

\twocolumn[{
    \renewcommand\twocolumn[1][]{#1}
    \maketitle    
    \begin{center}
        \vspace{-10mm}
        \includegraphics[width=0.75\textwidth]{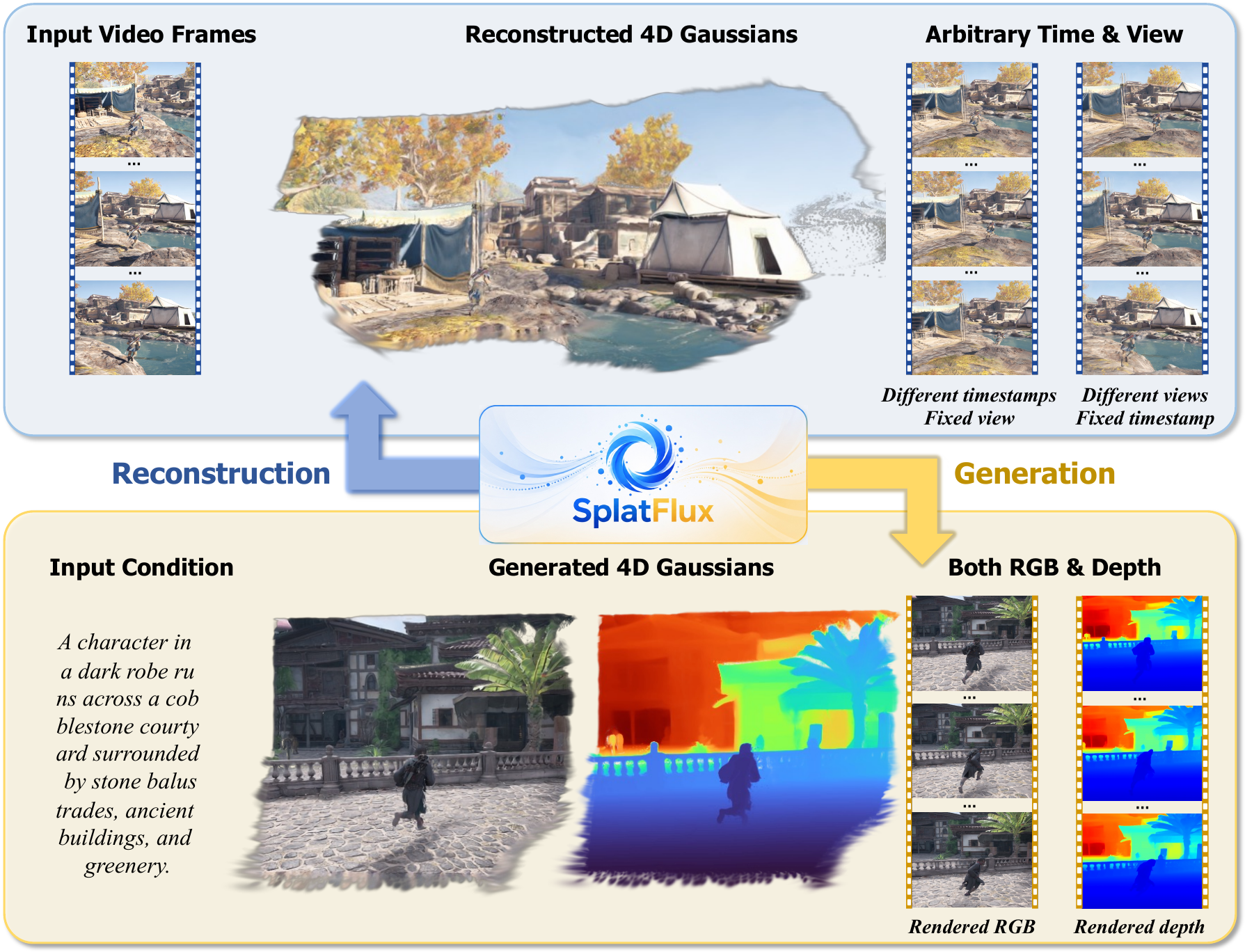}
        \captionsetup{hypcap=false,skip=4pt}
        \captionof{figure}{SplatFlux provides a feed-forward latent-to-4DGS framework for 4D reconstruction and generation, supporting arbitrary-time and arbitrary-view rendering with a shared dynamic Gaussian representation. It reconstructs 4D Gaussians from input video frames and generates 4D Gaussians from input conditions, producing coherent RGB and depth sequences.}
        \label{fig:teaser}
        \vspace{-2mm}
    \end{center}
}]

\begingroup
    \renewcommand{\thefootnote}{\fnsymbol{footnote}}
    \footnotetext[1]{Equal contribution.}
    \footnotetext[2]{Internship at Ant Group.}
    \footnotetext[3]{Corresponding authors.}
\endgroup
\setcounter{footnote}{0}

\begin{abstract}
This work presents \textbf{SplatFlux}, a \textbf{feed-forward latent-to-4DGS framework} for both video-based 4D reconstruction and conditional 4D generation. Our key observation is that observed videos and conditionally generated content can be represented in the same pretrained video latent space, allowing both tasks to be formulated as latent-to-4DGS decoding. To construct compact and temporally coherent 4D scenes, SplatFlux introduces a static-dynamic \textbf{decoupled 4D Gaussian representation} that explicitly predicts dynamic probabilities for Gaussian primitives. This representation separates static scene structures from moving regions, enabling complete 4D scene construction in a feed-forward manner through static aggregation and dynamic motion interpolation. To learn static-dynamic decomposition and dynamic Gaussian motion without costly annotations, we further develop an \textbf{auxiliary distillation strategy} that provides pseudo dynamic masks and optical-flow cues. Extensive experiments across 3D and 4D reconstruction and generation demonstrate that SplatFlux achieves state-of-the-art performance, producing high-quality dynamic scenes through a unified latent-to-4DGS framework.
\end{abstract}

\section{Introduction}
Reconstructing and synthesizing dynamic 3D worlds is essential for filmmaking, VR/AR, and physical AI systems grounded in closed-loop simulation~\citep{Lyra, Diff4Splat, Aether, CAT4D, 4Real-Video, DynamicVerse}. Recent feed-forward geometry and depth models~\citep{VGGT, DA3, Pi3, CUT3R, DUSt3R} demonstrate that rich 3D information can be inferred directly from visual observations. Building on this progress, feed-forward 4D Gaussian Splatting (4DGS) methods~\citep{4DGT, NeoVerse, BTimer, MoVieS, UFO-4D} have emerged to predict temporally coherent Gaussian primitives for efficient dynamic scene reconstruction and novel view synthesis.

Despite this progress, most existing feed-forward 4DGS approaches~\citep{4DGT, MoVieS, NeoVerse, BTimer, UFO-4D, StreamSplat} are developed for reconstructing dynamic scenes from observed videos. In parallel, modern video generative models~\citep{Mochi, HunyuanVideo, Wan, CogVideoX} can synthesize high-quality video latents from user conditions, providing conditionally generated dynamic visual content in the video latent domain. These advances suggest a latent formulation of 4DGS, where both captured videos and generated content can be represented as video latent sequences before being decoded into dynamic Gaussian representations.

Nevertheless, a key challenge lies in separating and representing static and dynamic components within a complete 4D scene. Existing feed-forward 4DGS approaches often separate static and dynamic components indirectly, through temporal lifespan modeling~\citep{4DGT}, tracking-based post-processing~\citep{NeoVerse}, or motion-induced displacement prediction~\citep{MoVieS}. Such implicit formulations may entangle moving objects with static scene structures, especially under complex motion and occlusion.

In this work, we present \textbf{SplatFlux}, a \textbf{feed-forward latent-to-4DGS framework} for both video-based reconstruction and conditional 4D generation. Our key observation is that observed videos and conditionally generated content can be represented in the same pretrained video latent space~\citep{Wan}. Observed videos are compressed into latent sequences by the VAE encoder, while conditional generation directly produces latent sequences in this space. This common latent representation allows both tasks to be formulated as a unified latent-to-4DGS decoding problem, enabling a shared decoder to map video latents from different sources into dynamic Gaussian representations.

Beyond this task-level unification, SplatFlux further addresses the representation-level ambiguity between static scene structures and moving regions. Instead of inducing static-dynamic separation implicitly from temporal cues, we introduce a static-dynamic \textbf{decoupled 4D Gaussian representation} by explicitly predicting dynamic probabilities for Gaussian primitives. This explicit representation enables complete 4D scene construction in a feed-forward manner without relying on hand-crafted post-processing or per-timestamp network inference. To learn static-dynamic decomposition and dynamic Gaussian motion without costly annotations, we adopt an \textbf{auxiliary distillation strategy}. Specifically, we supervise dynamic probabilities with pseudo dynamic masks generated by vision-language-guided segmentation~\citep{Qwen2.5-VL, Sa2VA}, and constrain dynamic Gaussian motion using pseudo optical-flow labels~\citep{UFM}. These complementary signals provide scalable supervision for explicit static-dynamic decomposition and motion learning.

We evaluate SplatFlux across 3D and 4D reconstruction and generation tasks. Experiments demonstrate that SplatFlux achieves state-of-the-art results. 

Our contributions are summarized as follows:

\begin{itemize}
    \item We propose SplatFlux, a feed-forward latent-to-4DGS framework that decodes video latents from different sources into 4D Gaussians, supporting both video-based 4D reconstruction and conditional 4D generation.

    \item We introduce a static-dynamic decoupled 4D Gaussian representation through explicit dynamic probability estimation, enabling compact and complete 4D scene construction in a feed-forward manner.

    \item We develop an auxiliary distillation strategy that uses vision-language-guided dynamic masks and optical-flow cues to learn static-dynamic decomposition and dynamic Gaussian motion without costly annotations.

    \item Extensive experiments across 3D and 4D reconstruction and generation tasks demonstrate that SplatFlux achieves state-of-the-art performance.
\end{itemize}

\section{Related Work}

\subsection{3D Reconstruction and Novel View Synthesis}
Traditional 3D reconstruction approaches typically rely on costly per-scene optimization, including classical structure-from-motion (SfM)~\citep{SfM}, neural radiance fields (NeRF)~\citep{NeRF}, and 3D Gaussian Splatting (3DGS)~\citep{3DGS}. Several methods~\citep{Shape_of_motion, Splatter_a_video, RetimeGS, MoSca} further extend 3DGS to dynamic scenes through temporal deformation modeling and additional supervision estimated from vision foundation models. With the growth of large-scale 3D datasets and foundation models~\citep{ViT, DINOv2, DINOv3, DPT}, feed-forward 3D and 4D reconstruction approaches~\citep{DUSt3R, MASt3R, MoGe, Pi3, DA3, VGGT, Dens3R, MoRE, LDCM, MonST3R, CUT3R, Easi3R, 4D-VGGT, VGGT4D, DynamicVerse, VGGT-omega, ViGeo, IGGT4D, MORE_4D} have become increasingly prominent. Methods such as DUSt3R~\citep{DUSt3R} and VGGT~\citep{VGGT} directly predict dense geometry and camera parameters from images, while approaches~\citep{YoNoSplat, NoPoSplat, WorldMirror, FLARE, C3G, Uni3R, PreF3R, Anchor3DGS} like AnySplat~\citep{AnySplat} predict 3D Gaussian primitives for novel view synthesis.

For feed-forward dynamic novel view synthesis, recent methods such as L4GM~\citep{L4GM}, STORM~\citep{STORM}, 4DGT~\citep{4DGT}, MoVieS~\citep{MoVieS}, DGS-LRM~\citep{DGS-LRM}, and BTimer~\citep{BTimer} estimate motion-related attributes for Gaussian primitives. However, most of these approaches still assume posed inputs, and MoVieS~\citep{MoVieS} and BTimer~\citep{BTimer} further require additional network inference for rendering at each target timestamp. UFO-4D~\citep{UFO-4D} relaxes the pose requirement, but is limited to two input views. C4G~\citep{C4G} adopts a compact Gaussian representation, but the limited number of Gaussians may constrain rendering quality. NeoVerse~\citep{NeoVerse} supports unposed video inputs, yet relies on hand-crafted global motion tracking to aggregate a scene-level representation. These limitations highlight the need for a feed-forward framework that can infer a complete dynamic Gaussian representation from unposed monocular videos, without per-timestamp network inference or hand-crafted post-processing.

\subsection{3D and 4D Generation}
Recent 3D and 4D generative approaches have made notable progress by distilling or fine-tuning pretrained image and video diffusion models~\citep{Imagen, StableDiffusion, Wan, CogVideoX}. The pioneering work DreamFusion~\citep{DreamFusion} introduces score distillation sampling (SDS) to optimize 3D representations by distilling gradients from pretrained diffusion models. Subsequent methods~\citep{ProlificDreamer, MAV3D, TC4D, 4D-fy} improve the distillation objective or extend this paradigm to 4D scene generation.

Another line of work~\citep{LGM, Bolt3D, CAT3D, DimensionX, SplatFlow, SEVA, Prometheus, DiffusionGS, GAE, PixWorld, GSExplorer} combines generative diffusion priors with 3D reconstruction for efficient 3D scene generation. Approaches such as CAT3D~\citep{CAT3D} and SEVA~\citep{SEVA} first synthesize multi-view images with diffusion models, followed by 3D reconstruction. In contrast, other methods~\citep{Prometheus, DiffusionGS, SplatFlow, VIST3A} bypass explicit RGB multi-view synthesis, and directly decode latent features into 3D Gaussian primitives. To support 4D content generation, some approaches~\citep{4Real-Video, CAT4D, SV4D} generate synchronized multi-view videos, while others extend video generation to dynamic point maps~\citep{Aether, 4DNeX, SteerX, BeyondPixels, VGGRPO}. Recent methods~\citep{Lyra, Diff4Splat, AEON} further explore dynamic Gaussian generation. However, methods like Lyra~\citep{Lyra} and Diff4Splat~\citep{Diff4Splat} rely on target-time Gaussian prediction mechanisms, as in BTimer~\citep{BTimer} and MoVieS~\citep{MoVieS}, where rendering a new timestamp requires additional network inference to obtain the corresponding Gaussian state. In contrast, SplatFlux formulates conditional 4D generation as latent-to-4DGS decoding, and directly produces a complete 4DGS representation that supports continuous-time rendering without per-timestamp network inference.

\section{Method}
\subsection{Overview}
\begin{figure*}[!t] 
    \centering
    \includegraphics[width=0.98\linewidth]{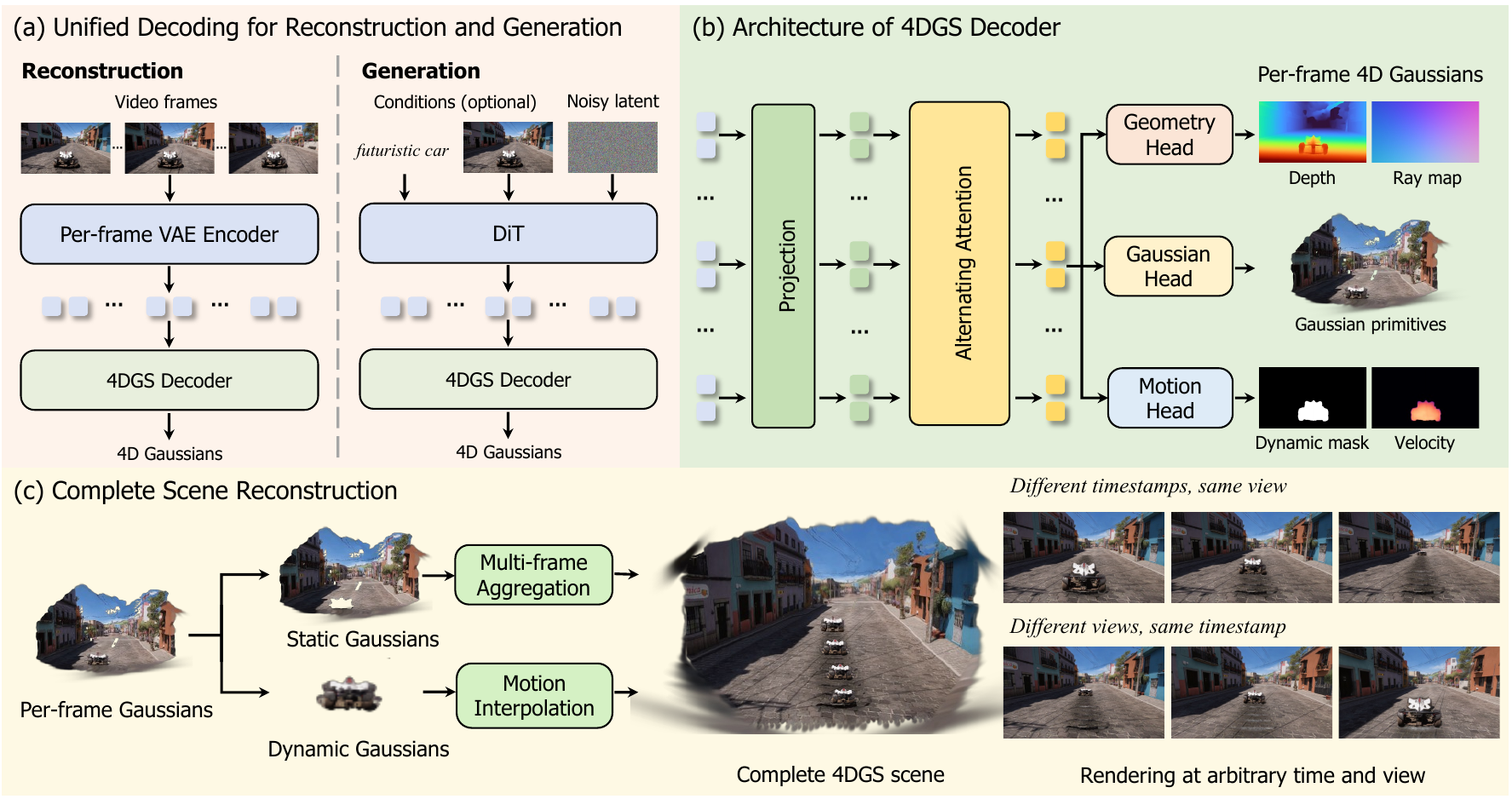} 
    \caption{Overview of SplatFlux. SplatFlux unifies video-based reconstruction and conditional generation through a shared latent-to-4DGS decoder. It predicts a static-dynamic decoupled 4D Gaussian representation, where static Gaussians are aggregated across frames and dynamic Gaussians are propagated by local motion for arbitrary-time and arbitrary-view rendering.}
    \label{fig:method}
\end{figure*}

We introduce SplatFlux, a feed-forward framework that unifies 4D Gaussian reconstruction and generation under a \textbf{latent-to-4DGS} formulation. 
Given a latent sequence $\mathbf{z}_{0:T-1}$\footnote{We use the pretrained Wan VAE latent space~\citep{Wan} as the common latent domain for both tasks.} obtained by encoding observed videos or by conditional generation, SplatFlux decodes it into a 4D Gaussian scene $\mathcal{S}$:
\begin{equation}
\mathcal{S} = \mathcal{D}_{\mathrm{GS}}(\mathbf{z}_{0:T-1}),
\end{equation}
where $\mathcal{D}_{\mathrm{GS}}$ is the shared 4DGS decoder. At the core of SplatFlux is a \textbf{static-dynamic decoupled 4D Gaussian representation}. Each Gaussian primitive is assigned a dynamic probability for explicit static-dynamic decomposition. Static Gaussians are aggregated across frames into a compact scene representation, while dynamic Gaussians remain frame-anchored and are propagated by local velocities for continuous-time rendering. This allows SplatFlux to construct a complete 4D Gaussian scene for arbitrary-view and arbitrary-time rendering without hand-crafted post-processing or per-timestamp network inference. Fig.~\ref{fig:method} illustrates the overall pipeline.

\subsection{Decoupled 4D Gaussian Representation}
\noindent\textbf{Per-frame 4DGS Representation.}
Given a latent sequence $\mathbf{z}_{0:T-1}$, our decoder predicts a set of per-pixel 4D Gaussian primitives for each timestamp. We denote the 4D Gaussians at frame $t$ as
\begin{equation}
\mathcal{G}_t=\left\{g_{t,i}
=
\left(
\boldsymbol{\mu}_{t,i},
\mathbf{s}_{t,i},
\mathbf{q}_{t,i},
\alpha_{t,i},
\mathbf{c}_{t,i},
m_{t,i},
\mathbf{v}^{+}_{t,i},
\mathbf{v}^{-}_{t,i}
\right)
\right\}_{i=1}^{HW},    
\end{equation}
where $i$ indexes the image grid, and $H$ and $W$ denote the output resolution. Each primitive $g_{t,i}$ is parameterized by its 3D center $\boldsymbol{\mu}_{t,i}$, scale $\mathbf{s}_{t,i}$, rotation $\mathbf{q}_{t,i}$, opacity $\alpha_{t,i}$, color $\mathbf{c}_{t,i}$, dynamic probability $m_{t,i}$, forward velocity $\mathbf{v}^{+}_{t,i}$, and backward velocity $\mathbf{v}^{-}_{t,i}$. We note that the dynamic probability enables explicit static-dynamic decomposition of Gaussian primitives in a feed-forward manner, while $\mathbf{v}^{+}_{t,i}$ and $\mathbf{v}^{-}_{t,i}$ support arbitrary-time rendering by locally interpolating Gaussian motion without per-timestamp network inference. 

\noindent\textbf{Complete Scene Reconstruction.}
To obtain a compact and complete 4DGS scene, we do not simply concatenate per-frame Gaussians across timestamps, as this would introduce substantial redundancy in static regions and entangle static and dynamic content. Instead, we explicitly separate the predicted Gaussians according to their dynamic probabilities. Given a threshold $\tau_m$, the Gaussians at frame $t$ are divided into static and dynamic subsets as
\begin{equation}
\begin{aligned}
\mathcal{G}^{s}_t&=\{g_{t,i}\mid m_{t,i}<\tau_m\}, \\ \mathcal{G}^{d}_t&=\{g_{t,i}\mid m_{t,i}\geq\tau_m\}.
\end{aligned}
\end{equation}
For static regions, we perform multi-frame aggregation through differentiable voxelization~\citep{AnySplat} to merge redundant primitives, obtaining $\mathcal{G}^{s}=\mathrm{Agg}\left(\bigcup_{t=0}^{T-1}\mathcal{G}^{s}_t\right)$. The complete 4DGS scene is therefore composed as $
\mathcal{S}=\mathcal{G}^{s}\cup \left\{\mathcal{G}^{d}_t\right\}_{t=0}^{T-1}$, where $\mathcal{G}^{s}$ represents the compact static scene, and $\{\mathcal{G}^{d}_t\}$ preserves dynamic regions with their associated velocities for continuous-time rendering. Fig.~\ref{fig:method}(c) illustrates the process of our complete scene reconstruction. 

For a target time $t_\mathrm{target}$ between two adjacent timestamps $t$ and $t+1$, we move the dynamic Gaussians $\mathcal{G}^{d}_{t}$ and $\mathcal{G}^{d}_{t+1}$ to $t_\mathrm{target}$ using linear motion interpolation~\citep{UFO-4D, NeoVerse, StreamSplat} as 
\begin{equation}
\begin{aligned}
\boldsymbol{\mu}_{t\rightarrow t_\mathrm{target},i} &= \boldsymbol{\mu}_{t,i} + (t_\mathrm{target}-t) \mathbf{v}^{+}_{t,i}, \\
\boldsymbol{\mu}_{t+1\rightarrow t_\mathrm{target},i} &= \boldsymbol{\mu}_{t+1,i} +
(t + 1 - t_\mathrm{target})\mathbf{v}^{-}_{t+1,i}.
\end{aligned}
\end{equation}
The scene at $t_\mathrm{target}$ is then composed of the static Gaussians and moved dynamic Gaussians, enabling rasterization~\citep{3DGS} of novel views. Overall, this static-dynamic decoupled representation provides a compact and complete 4D scene by aggregating static structures across frames while preserving dynamic regions with local motion.

\subsection{Unified Latent-to-4DGS Decoding}
\noindent\textbf{Reconstruction from Videos.} Given a monocular video sequence $\mathbf{I}_{0:T-1}=\{I_0,\ldots,I_{T-1}\}$, we use the frozen Wan VAE encoder $\mathcal{E}$ to extract latents as $ \mathbf{z}_t = \mathcal{E}(I_t)$. We note that each frame is treated as an individual image and encoded independently, without applying temporal compression in the video VAE encoder. This frame-wise design avoids temporal compression, and allows training with unordered multi-view image datasets.

\noindent\textbf{Generation from Conditions.} For input conditions $c$, such as a text prompt, a single image, or their combination, we use a Diffusion Transformer (DiT)~\citep{DiT} as the generator $\Phi_{\mathrm{gen}}$ to produce a latent sequence as $\mathbf{z}_{0:T-1} = \Phi_{\mathrm{gen}}(c)$. Different from standard video generation pipelines that decode generated latents into RGB videos, we directly decode the generated latent sequence into a 4DGS scene using our proposed decoder $\mathcal{D}_{\textrm{GS}}$.

\noindent\textbf{Architecture of 4DGS Decoder.}
As shown in Fig.~\ref{fig:method}(b), the decoder first maps the low-dimensional VAE latent sequence into high-dimensional feature tokens through a latent projector. The resulting tokens are then processed by alternating attention blocks~\citep{VGGT} to exchange spatial and temporal information. Finally, we use three prediction heads~\citep{DPT} to estimate geometry, appearance, and motion attributes from the spatiotemporal fused features $\mathbf{F}_t$ as
\begin{equation}
\begin{aligned}
(\mathbf{d}_t,\mathbf{r}_t)&=\mathcal{H}_{\mathrm{geo}}(\mathbf{F}_t), \\
\mathbf{a}_t&=\mathcal{H}_{\mathrm{app}}(\mathbf{F}_t), \\
(\mathbf{m}_t,\mathbf{v}^{+}_t,\mathbf{v}^{-}_t)&=\mathcal{H}_{\mathrm{motion}}(\mathbf{F}_t),
\end{aligned}
\end{equation}
where $\mathbf{d}_t$ and $\mathbf{r}_t$ denote depth and ray map~\citep{DA3}, $\mathbf{a}_t$ denotes appearance attributes including scale, rotation, opacity, and color, and $\mathbf{m}_t$, $\mathbf{v}^{+}_t$, and $\mathbf{v}^{-}_t$ denote the dynamic mask, forward velocity, and backward velocity, respectively. The Gaussian centers are derived from the predicted $\mathbf{d}_t$ and $\mathbf{r}_t$.

\subsection{Model Training}
Learning static-dynamic decomposition and dynamic Gaussian motion would benefit from direct supervision with dynamic masks and motion labels. However, such annotations are costly to obtain, and are typically limited to small-scale synthetic datasets~\citep{dynamicreplica, pointodyssey}. Relying only on rendering losses provides weak constraints on dynamic probabilities and local velocities, often leading to ambiguous decomposition and inaccurate motion estimation.

To address this issue, we adopt an \textbf{auxiliary distillation strategy}. We generate pseudo dynamic masks and optical-flow labels using expert foundation models~\citep{Qwen2.5-VL, Sa2VA, UFM}, and use them as auxiliary supervision during training. This provides scalable guidance on where the scene should be decomposed and how dynamic Gaussians should move, allowing SplatFlux to benefit from large-scale training without costly annotations.

\noindent\textbf{Dynamic Mask Supervision.} 
To construct pseudo labels for dynamic masks, we first use Qwen2.5-VL-72B-Instruct~\citep{Qwen2.5-VL} to detect dynamic objects in videos, and generate a text description including the category and visual attributes for each detected dynamic object. 
We then use this description as the prompt for Sa2VA~\citep{Sa2VA} to segment the corresponding object. 
The masks of all detected dynamic objects in the same image are merged into a pseudo dynamic mask.

Given the predicted dynamic mask $\mathbf{m}_t$ and its pseudo label $\widetilde{\mathbf{m}}_t$, we compute the dynamic mask loss as a combination of binary cross-entropy, Dice, and binarization regularization:
\begin{equation}
\begin{aligned}
\mathcal{L}_{\mathrm{dyn}}
={}&\mathcal{L}_{\mathrm{BCE}}(\mathbf{m}_t, \widetilde{\mathbf{m}}_t) \\
&+\lambda_{\mathrm{Dice}}\mathcal{L}_{\mathrm{Dice}}(\mathbf{m}_t, \widetilde{\mathbf{m}}_t) \\
&+\lambda_{\mathrm{bin}}\mathcal{L}_{\mathrm{bin}}(\mathbf{m}_t),
\end{aligned}
\end{equation}
where the binarization regularization is defined as
\begin{equation}
\mathcal{L}_{\mathrm{bin}}(\mathbf{m}_t)
=
\frac{1}{|\Omega|}
\sum_{\mathbf{x}\in\Omega}
\mathbf{m}_t(\mathbf{x})
\left(1-\mathbf{m}_t(\mathbf{x})\right),
\end{equation}
and $\Omega$ denotes all image pixels. 
The dynamic mask supervision provides an explicit signal for separating static and dynamic Gaussians, which is essential for feed-forward 4D scene construction.

\noindent\textbf{Motion Supervision.} 
To supervise local velocities, we employ the optical-flow foundation model UFM~\citep{UFM} to generate pseudo motion cues. 
Instead of directly supervising velocities in 3D space, we project the predicted Gaussian motion into the image plane, and minimize the discrepancy between the optical flow induced by the predicted Gaussian motion and the pseudo flow labels from UFM.

Given the predicted depth $\mathbf{d}_t$, ray map $\mathbf{r}_t$, and bidirectional velocities $\mathbf{v}^{+}_t$ and $\mathbf{v}^{-}_t$, we compute the Gaussian center at pixel $\mathbf{x}\in\Omega$ as $ \boldsymbol{\mu}_t(\mathbf{x}) = \mathcal{P}\bigl(\mathbf{x}, \mathbf{d}_t(\mathbf{x}), \mathbf{r}_t(\mathbf{x})\bigr)$, where $\mathcal{P}(\cdot)$ denotes the backward projection function. 
The optical flows from frame $t$ to $t+1$ and $t-1$ can be derived as
\begin{equation}
\mathbf{F}_{t\rightarrow t+1}(\mathbf{x}) =
\Pi_{t+1} \left( \boldsymbol{\mu}_t(\mathbf{x}) + \mathbf{v}^{+}_t(\mathbf{x}) \right) - \mathbf{x},
\end{equation}
\begin{equation}
\mathbf{F}_{t\rightarrow t-1}(\mathbf{x}) = 
\Pi_{t-1} \left( \boldsymbol{\mu}_t(\mathbf{x}) + \mathbf{v}^{-}_t(\mathbf{x}) \right) - \mathbf{x}.
\end{equation}
Here, $\Pi_{t+1}(\cdot)$ and $\Pi_{t-1}(\cdot)$ denote the projection functions using the corresponding ray maps. 
Let $\widetilde{\mathbf{F}}_{t\rightarrow t+1}$ and 
$\widetilde{\mathbf{F}}_{t\rightarrow t-1}$ denote the pseudo optical flows estimated by the teacher model. 
We supervise the optical flows induced by Gaussian motion as
\begin{equation}
\begin{aligned}
\mathcal{L}_{\mathrm{flow}}
={}& \frac{1}{|\Omega|} \sum_{\mathbf{x}\in\Omega} w_t(\mathbf{x}) \\
&\quad\Bigl(\left\| \mathbf{F}_{t\rightarrow t+1}(\mathbf{x}) - \widetilde{\mathbf{F}}_{t\rightarrow t+1}(\mathbf{x}) \right\|_1 \\
&\qquad+\left\| \mathbf{F}_{t\rightarrow t-1}(\mathbf{x}) - \widetilde{\mathbf{F}}_{t\rightarrow t-1}(\mathbf{x}) \right\|_1\Bigr),
\end{aligned}
\end{equation}
where $w_t(\mathbf{x})$ is a dynamic-region weight derived from the pseudo dynamic mask $\widetilde{\mathbf{m}}_t$. 
This weighting encourages the model to focus more on regions with object motion.

\noindent\textbf{Overall Training Objective.}
The overall training objective combines geometry, rendering, dynamic mask, and optical-flow losses as
\begin{equation}
\label{eq:loss_gs}
\mathcal{L} = 
\lambda_{\mathrm{geo}}\mathcal{L}_{\mathrm{geo}} 
+ \lambda_{\mathrm{rend}}\mathcal{L}_{\mathrm{rend}} 
+ \lambda_{\mathrm{dyn}}\mathcal{L}_{\mathrm{dyn}} 
+ \lambda_{\mathrm{flow}}\mathcal{L}_{\mathrm{flow}}.
\end{equation}
Here, $\mathcal{L}_{\mathrm{geo}}$ supervises the depths and ray maps of source views following the implementation of DA3~\citep{DA3}. The rendering loss $\mathcal{L}_{\mathrm{rend}}$ is applied to the rendered RGB images and depths of both source and target views. Specifically, we use $\ell_2$ and LPIPS~\citep{LPIPS} losses for RGB rendering, and $\ell_1$ loss for depth rendering. The dynamic mask loss $\mathcal{L}_{\mathrm{dyn}}$ provides direct supervision for static-dynamic decomposition, while the optical-flow loss $\mathcal{L}_{\mathrm{flow}}$ constrains the motion of dynamic Gaussians.

\subsection{Implementation Details}
\label{sec:implementation}

\noindent\textbf{Training Datasets.} We train SplatFlux on 17 public datasets, covering static and dynamic content across object-centric, indoor, and outdoor scenarios. Table~\ref{tab:training_datasets} summarizes the 17 datasets used for training SplatFlux. We mainly follow CUT3R~\citep{CUT3R} for data preprocessing and filtering. For large-scale datasets such as Aria Synthetic Environments~\citep{ase} and Lyra~\citep{Lyra}, we select 5000 scenes for efficient training. For datasets with provided dynamic masks, such as OmniWorld-Game~\citep{omniworld} and Dynamic Replica~\citep{dynamicreplica}, we directly use the masks provided in the dataset. For the remaining dynamic datasets, such as Lyra~\citep{Lyra} and Stereo4D~\citep{stereo4d}, we generate pseudo dynamic masks using Qwen2.5-VL-72B-Instruct~\citep{Qwen2.5-VL} and Sa2VA~\citep{Sa2VA}. During training, we use UFM~\citep{UFM} to compute pseudo optical-flow supervision for all dynamic datasets. All 17 datasets are used to train the 4DGS decoder, while the generator is trained on DL3DV~\citep{dl3dv}, OmniWorld-Game~\citep{omniworld}, Stereo4D~\citep{stereo4d}, and Lyra~\citep{Lyra}.

\begin{table}[!t]
    \centering
    \caption{An overview of the training datasets used by SplatFlux.}
    \label{tab:training_datasets}
    \small
    \resizebox{\linewidth}{!}{\setlength{\tabcolsep}{3pt}
     \renewcommand{\arraystretch}{1.08}   
        \begin{tabular}{@{}lcccr@{}}
            \toprule
            Dataset & Domain & Real? & Dynamic? & \#Scenes \\
            \midrule
            ARKitScenes~\citep{arkitscenes} & Indoor & \cmark & \xmark & 675 \\
            BEDLAM~\citep{bedlam} & Mixed & \xmark & \cmark & 7309 \\
            CO3Dv2~\citep{co3dv2} & Object & \cmark & \xmark & 2500 \\
            DL3DV~\citep{dl3dv} & Mixed & \cmark & \xmark & 6370 \\
            Dynamic Replica~\citep{dynamicreplica} & Indoor & \xmark & \cmark & 483 \\
            Hypersim~\citep{hypersim} & Indoor & \xmark & \xmark & 741 \\
            Lyra~\citep{Lyra} & Mixed & \xmark & \cmark & 5000 \\
            ASE~\citep{ase} & Indoor & \xmark & \xmark & 5000 \\
            MVS-Synth~\citep{mvssynth} & Outdoor & \xmark & \cmark & 120 \\
            OmniWorld-Game~\citep{omniworld} & Mixed & \xmark & \cmark & 11424 \\
            PointOdyssey~\citep{pointodyssey} & Mixed & \xmark & \cmark & 33 \\
            ScanNet++~\citep{scannetpp} & Indoor & \cmark & \xmark & 853 \\
            Spring~\citep{spring} & Mixed & \xmark & \cmark & 35 \\
            Stereo4D~\citep{stereo4d} & Mixed & \cmark & \cmark & 5000 \\
            TartanAir~\citep{tartanair} & Mixed & \xmark & \cmark & 369 \\
            VKITTI2~\citep{vkitt2} & Outdoor & \xmark & \cmark & 100 \\
            Waymo~\citep{waymo} & Outdoor & \cmark & \cmark & 798 \\
            \bottomrule
        \end{tabular}
    }
\end{table}

\noindent\textbf{Training of the 4DGS Decoder.} We initialize the alternating attention blocks and geometry head from DA3-Giant~\citep{DA3}, use the frozen Wan2.1 VAE~\citep{Wan} as the latent encoder, and adopt gsplat~\citep{gsplat} for differentiable rasterization. For each training iteration, we sample $T$ source frames and $T-1$ intermediate target frames between adjacent source frames. The 4DGS decoder is supervised using the objective in Eq.~\ref{eq:loss_gs}. To improve generalization, we randomly sample the aspect ratio from $0.5$ to $2.0$, and resize and crop images while keeping the total number of pixels between $100{,}000$ and $250{,}000$. The number of source frames $T$ is randomly sampled from $2$ to $12$. Following previous works~\citep{VGGT, DA3}, we normalize the scene scale during training. We set the dynamic probability threshold to $\tau_m=0.05$ and the voxel size for static Gaussian aggregation to $0.001$ in the normalized scene coordinate system. We train the 4DGS decoder in two stages. In the first stage, we disable voxelization-based aggregation and train the decoder for $100$K iterations. In the second stage, we enable voxelization-based aggregation to reduce redundant static Gaussians and fine-tune the model for another $50$K iterations. We use the AdamW optimizer with a cosine learning rate schedule and linear warm-up. The maximum learning rates of the two stages are set to $1\times10^{-4}$ and $1\times10^{-5}$, respectively. For the parameters initialized from DA3-Giant~\citep{DA3}, we scale the learning rate by $0.1$.

\noindent\textbf{Training of the DiT Generator.} For conditional generation, we fine-tune Wan2.1-14B~\citep{Wan} with the standard flow-matching objective, where training videos are encoded into per-frame latent sequences without temporal compression. The generated conditional latents are directly decoded by the shared 4DGS decoder $\mathcal{D}_{\mathrm{GS}}$. We train the DiT generator to produce key frame latent sequences. The generator is fine-tuned for $20{,}000$ iterations using LoRA~\citep{LoRA}, with rank $r=32$, scaling factor $\alpha=64$, and group size $G=64$. We optimize the LoRA parameters with AdamW using a learning rate of $1\times10^{-4}$ and a weight decay of $1\times10^{-4}$.

The full training pipeline takes approximately 10 days on 32 NVIDIA H20 GPUs.

\section{Experiments}
This section evaluates SplatFlux on dynamic novel view synthesis, video depth estimation, static novel view synthesis, dynamic object segmentation, and conditional 3D and 4D generation, together with ablation studies and qualitative analyses.

\subsection{Dynamic Novel View Synthesis}
\noindent\textbf{Evaluation Protocol.}
For NVIDIA Dynamic Scene, we select 12 input frames in a round-robin manner across cameras, and test on 12 frames from camera 0. For DyCheck, we sample 64 consecutive timestamps for each scene. We reconstruct from 32 frames sampled every two timestamps from a casually captured video (camera 0), and evaluate on all 64 frames from a fixed-camera video (camera 1) together with the remaining 32 frames from camera 0. For DAVIS, we select 60 scenes for evaluation. For each scene, we sample 48 consecutive frames and split them evenly into 24 input frames and 24 evaluation frames. The image resolutions for NVIDIA Dynamic Scene, DyCheck, and DAVIS are set to $560 \times 336$, $336 \times 448$, and $560 \times 336$, respectively.

At evaluation time, we initialize the target camera poses through two forward passes~\citep{AnySplat,NeoVerse}. The first pass takes only the source views to reconstruct the Gaussian scene and estimate the source cameras. The second pass takes both source and target views to estimate cameras for the complete sequence. Using the shared source views, we align the two sets of camera poses and transform the target poses into the reconstructed scene's coordinate system. We then refine the initialized target poses via gradient descent on the photometric error between rendered and target images, while keeping the Gaussian scene fixed~\citep{NoPoSplat,YoNoSplat}. This refinement is used only for evaluation, and all methods use the same source and target views.

\begin{table*}[!t]
    \centering
    \caption{\textbf{Quantitative comparison of dynamic novel view synthesis} on the NVIDIA Dynamic Scene, DyCheck, and DAVIS datasets. * denotes approaches that require camera pose as input. The \textbf{best} and \underline{second-best} results are highlighted for clarity.}
    \label{tab:cmp_dyn_nvs}
    \resizebox{0.93\textwidth}{!}
    {
        \setlength{\tabcolsep}{6pt}
        \renewcommand{\arraystretch}{1.1}
        \begin{tabular}{lccccccccc}
            \toprule
            \multirow{2}{*}{\textbf{Method}} &
            \multicolumn{3}{c}{\textbf{NVIDIA Dynamic Scene}} &
            \multicolumn{3}{c}{\textbf{DyCheck}} &
            \multicolumn{3}{c}{\textbf{DAVIS}} \\
            \cmidrule(lr){2-4} \cmidrule(lr){5-7} \cmidrule(lr){8-10}
            & PSNR $\uparrow$ & SSIM $\uparrow$ & LPIPS $\downarrow$
            & mPSNR $\uparrow$ & mSSIM $\uparrow$ & mLPIPS $\downarrow$
            & PSNR $\uparrow$ & SSIM $\uparrow$ & LPIPS $\downarrow$ \\
            \midrule
            4DGT*~\citep{4DGT}
            & 16.57 & 0.359 & 0.591
            & 15.60 & 0.519 & 0.576
            & 16.64 & 0.459 & 0.589 \\
            MoVieS*~\citep{MoVieS}
            & 19.76 & \underline{0.606} & \underline{0.255}
            & \underline{16.52} & \underline{0.566} & \underline{0.403}
            & \underline{21.34} & \underline{0.698} & \underline{0.234} \\
            NeoVerse~\citep{NeoVerse}
            & \underline{20.20} & 0.582 & 0.280
            & 16.48 & 0.524 & 0.436
            & 19.31 & 0.564 & 0.333 \\
            C4G~\citep{C4G}
            & 15.43 & 0.351 & 0.675
            & 14.77 & 0.495 & 0.609
            & 14.90 & 0.394 & 0.712 \\
            AEON~\citep{AEON}
            & 19.59 & 0.468 & 0.404
            & 15.85 & 0.509 & 0.494
            & 21.04 & 0.609 & 0.331 \\
            \midrule
            SplatFlux (Ours)
            & \textbf{23.03} & \textbf{0.685} & \textbf{0.205}
            & \textbf{17.58} & \textbf{0.576} & \textbf{0.387}
            & \textbf{22.71} & \textbf{0.719} & \textbf{0.228} \\
            \bottomrule
        \end{tabular}
    }
\end{table*}

\begin{figure*}[!t] 
    \centering
    \includegraphics[width=0.95\linewidth]{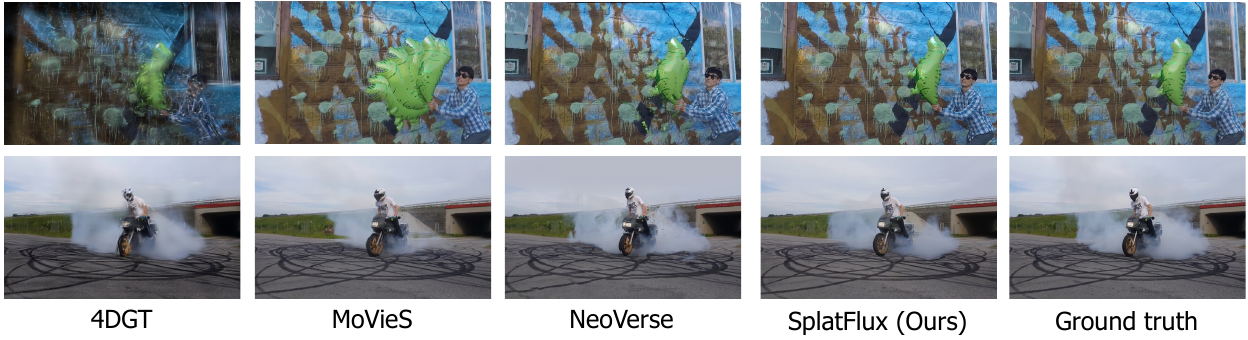}
    \caption{Qualitative comparison of dynamic novel view synthesis across different methods.}
    \label{fig:dyn_render}
\end{figure*}

\noindent\textbf{Comparison with State-of-the-Art Methods.}
We evaluate SplatFlux for dynamic novel view synthesis on the NVIDIA Dynamic Scene~\citep{NVIDIA_Dynamic_Scene}, DyCheck~\citep{Dycheck}, and DAVIS~\citep{DAVIS} datasets. We compare with state-of-the-art open-sourced methods~\citep{4DGT, MoVieS, NeoVerse, C4G, AEON}, and list the results in Table~\ref{tab:cmp_dyn_nvs}. The results demonstrate that SplatFlux achieves superior dynamic novel view synthesis without relying on camera poses as input. The qualitative comparison in Fig.~\ref{fig:dyn_render} further shows that our method produces clearer details with fewer rendering artifacts. In contrast, the comparison methods often suffer from ghosting artifacts in dynamic regions, or exhibit limited generalization to in-the-wild scenes.

\subsection{Video Depth Estimation}
We evaluate the video depth estimation capability of SplatFlux on the Sintel~\citep{sintel}, Bonn~\citep{bonn}, and KITTI~\citep{kitti} datasets. In line
with ViGeo~\citep{ViGeo}, the predicted depth sequences are aligned to the ground-truth depth using a scale factor per sequence. Table~\ref{tab:cmp_video_depth} reports the absolute relative error (Abs Rel) and threshold accuracy ($\delta_{1}$). SplatFlux achieves competitive performance, outperforming 3D reconstruction foundation models such as VGGT~\citep{VGGT} and DA3~\citep{DA3}.

\begin{table}[!t]
    \centering
    \caption{\textbf{Quantitative comparison of scale-invariant video depth estimation} on the Sintel~\citep{sintel}, Bonn~\citep{bonn}, and KITTI~\citep{kitti} datasets. The \textbf{best} and \underline{second-best} results are highlighted for clarity.}
    \label{tab:cmp_video_depth}
    \footnotesize
    \resizebox{\linewidth}{!}{\setlength{\tabcolsep}{2.5pt}
        \renewcommand{\arraystretch}{1.1}
        \begin{tabular}{@{}lcccccc@{}}
            \toprule
            \multirow{2}{*}{\textbf{Method}} &
            \multicolumn{2}{c}{\textbf{Sintel}} &
            \multicolumn{2}{c}{\textbf{Bonn}} &
            \multicolumn{2}{c}{\textbf{KITTI}} \\
            \cmidrule(lr){2-3} \cmidrule(lr){4-5} \cmidrule(lr){6-7}
            & Rel $\downarrow$ & $\delta_{1}$ $\uparrow$ & Rel $\downarrow$ & $\delta_{1}$ $\uparrow$ & Rel $\downarrow$ & $\delta_{1}$ $\uparrow$ \\
            \midrule            
            VDA~\citep{vda} & 0.855 & 0.340 & 0.301 & 0.465 & 0.272 & 0.535 \\
            DepthCrafter~\citep{DepthCrafter} & 0.469 & 0.455 & 0.074 & 0.968 & 0.282 & 0.435 \\
            GeometryCrafter~\citep{GeometryCrafter} & 0.305 & 0.675 & 0.057 & 0.969 & 0.057 & 0.981 \\
            VGGT~\citep{VGGT} & 0.373 & 0.634 & 0.057 & 0.966 & 0.062 & 0.969 \\
            Pi3~\citep{Pi3} & 0.281 & 0.663 & \underline{0.049} & \underline{0.975} & \underline{0.038} & \underline{0.985} \\
            DA3~\citep{DA3}  & 0.284 & 0.666 & 0.050 & 0.972 & 0.058 & 0.976 \\
            ViGeo~\citep{ViGeo} & \textbf{0.229} & \textbf{0.724} & \textbf{0.046} & \textbf{0.976} & \textbf{0.036} & 0.983 \\
            \midrule
            SplatFlux (Ours) & \underline{0.248} & \underline{0.679} & \underline{0.049} & 0.974 & \textbf{0.036} & \textbf{0.986} \\
            \bottomrule
        \end{tabular}
    }
\end{table}

\subsection{Static Novel View Synthesis}
SplatFlux also supports 3D Gaussian reconstruction and rendering of static scenes. We evaluate SplatFlux on the DL3DV-140 benchmark~\citep{dl3dv}, and report the results in Table~\ref{tab:cmp_static_nvs}. SplatFlux achieves strong overall performance, demonstrating its generalization to static novel view synthesis.

\begin{table}[!t]
    \centering
    \caption{\textbf{Quantitative comparison of static novel view synthesis} on the DL3DV-140 benchmark~\citep{dl3dv} with 6 and 12 unposed input views. The \textbf{best} and \underline{second-best} results are highlighted.}
    \label{tab:cmp_static_nvs}
    \footnotesize
    \resizebox{\linewidth}{!}{
        \setlength{\tabcolsep}{2.5pt}
        \renewcommand{\arraystretch}{1.1}
        \begin{tabular}{@{}lcccccc@{}}
            \toprule
            \multirow{2}{*}{\textbf{Method}} &
            \multicolumn{3}{c}{\textbf{6 Views}} &
            \multicolumn{3}{c}{\textbf{12 Views}} \\
            \cmidrule(lr){2-4} \cmidrule(lr){5-7} & PSNR $\uparrow$ & SSIM $\uparrow$ & LPIPS $\downarrow$ & PSNR $\uparrow$ & SSIM $\uparrow$ & LPIPS $\downarrow$ \\
            \midrule
            WorldMirror~\citep{WorldMirror} & \underline{21.69} & \underline{0.7050} & 0.2418 & \underline{20.61} & \underline{0.6558} & 0.2933 \\
            DA3~\citep{DA3} & 20.86 & 0.6472 & \underline{0.2395} & 20.54 & 0.6241 & \textbf{0.2592} \\
            NeoVerse~\citep{NeoVerse} & 20.02 & 0.6170 & 0.3267 & 18.99 & 0.5550 & 0.3915 \\
            \midrule
            SplatFlux (Ours) & \textbf{22.17} & \textbf{0.7309} & \textbf{0.2287} & \textbf{21.01} & \textbf{0.6731} & \underline{0.2778} \\
            \bottomrule
        \end{tabular}%
    }
\end{table}

\begin{table*}[!t]
    \centering
    \caption{\textbf{Quantitative comparison of dynamic object segmentation} on the DAVIS~\citep{DAVIS} dataset. Easi3R$_\textrm{DUSt3R/MonST3R}$ denotes the variant using DUSt3R~\citep{DUSt3R} or MonST3R~\citep{MonST3R} as the backbone, and w/ SAM2 denotes using SAM2~\citep{SAM2} for mask refinement. The \textbf{best} and \underline{second-best} results are highlighted for clarity.}
    \label{tab:cmp_dyn_mask}
    {
        \fontsize{8.75}{10.5}\selectfont   
        \setlength{\tabcolsep}{4pt}
        \renewcommand{\arraystretch}{1.1}
            \begin{tabular}{lcccccccccccc}
            \toprule
            \multirow{2}{*}{\textbf{Method}} &
            \multicolumn{4}{c}{\textbf{DAVIS-16}} &
            \multicolumn{4}{c}{\textbf{DAVIS-17}} &
            \multicolumn{4}{c}{\textbf{DAVIS-all}} \\
            \cmidrule(lr){2-5} \cmidrule(lr){6-9} \cmidrule(lr){10-13}
            & JM $\uparrow$ & JR $\uparrow$ & FM $\uparrow$ & FR $\uparrow$
            & JM $\uparrow$ & JR $\uparrow$ & FM $\uparrow$ & FR $\uparrow$
            & JM $\uparrow$ & JR $\uparrow$ & FM $\uparrow$ & FR $\uparrow$ \\
            \midrule
            Easi3R$_\textrm{DUSt3R}$ (w/ SAM2)~\citep{Easi3R} & \underline{71.7} & 80.5 & \underline{71.9} & 78.1 & 64.3 & 71.8 & 64.3 & 69.4 & 56.4 & 63.2 & 55.6 & 58.5 \\
            Easi3R$_\textrm{MonST3R}$ (w/ SAM2)~\citep{Easi3R} & 70.7 & \underline{81.8} & 68.4 & \underline{79.4} & \underline{66.9} & \underline{76.2} & \underline{65.2} & \underline{75.7} & \underline{63.2} & \underline{71.1} & \underline{61.0} & \underline{66.9} \\
            VGGT4D~\citep{VGGT4D} & 62.1 & 76.8 & 56.0 & 67.5 & 56.5 & 65.6 & 51.1 & 56.9 & 50.8 & 55.6 & 47.0 & 46.4 \\
            \midrule
            SplatFlux (Ours) & \textbf{76.8} & \textbf{89.3} & \textbf{81.2} & \textbf{92.5} & \textbf{76.7} & \textbf{88.1} & \textbf{80.9} & \textbf{91.2} & \textbf{73.6} & \textbf{82.5} & \textbf{78.6} & \textbf{87.3} \\
            \bottomrule
        \end{tabular}
    }
\end{table*}

\begin{figure*}[!t] 
    \centering
    \includegraphics[width=0.98\linewidth]{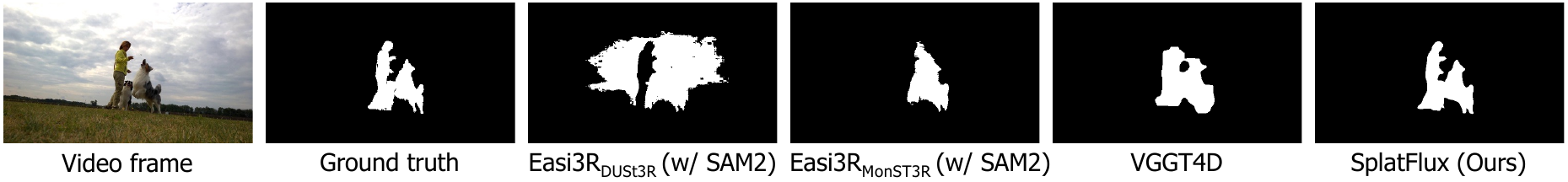} 
    \caption{Qualitative comparison of dynamic object segmentation across different methods.}
    \label{fig:dynmask}
\end{figure*}

\begin{figure*}[!t] 
    \centering
    \includegraphics[width=0.96\linewidth]{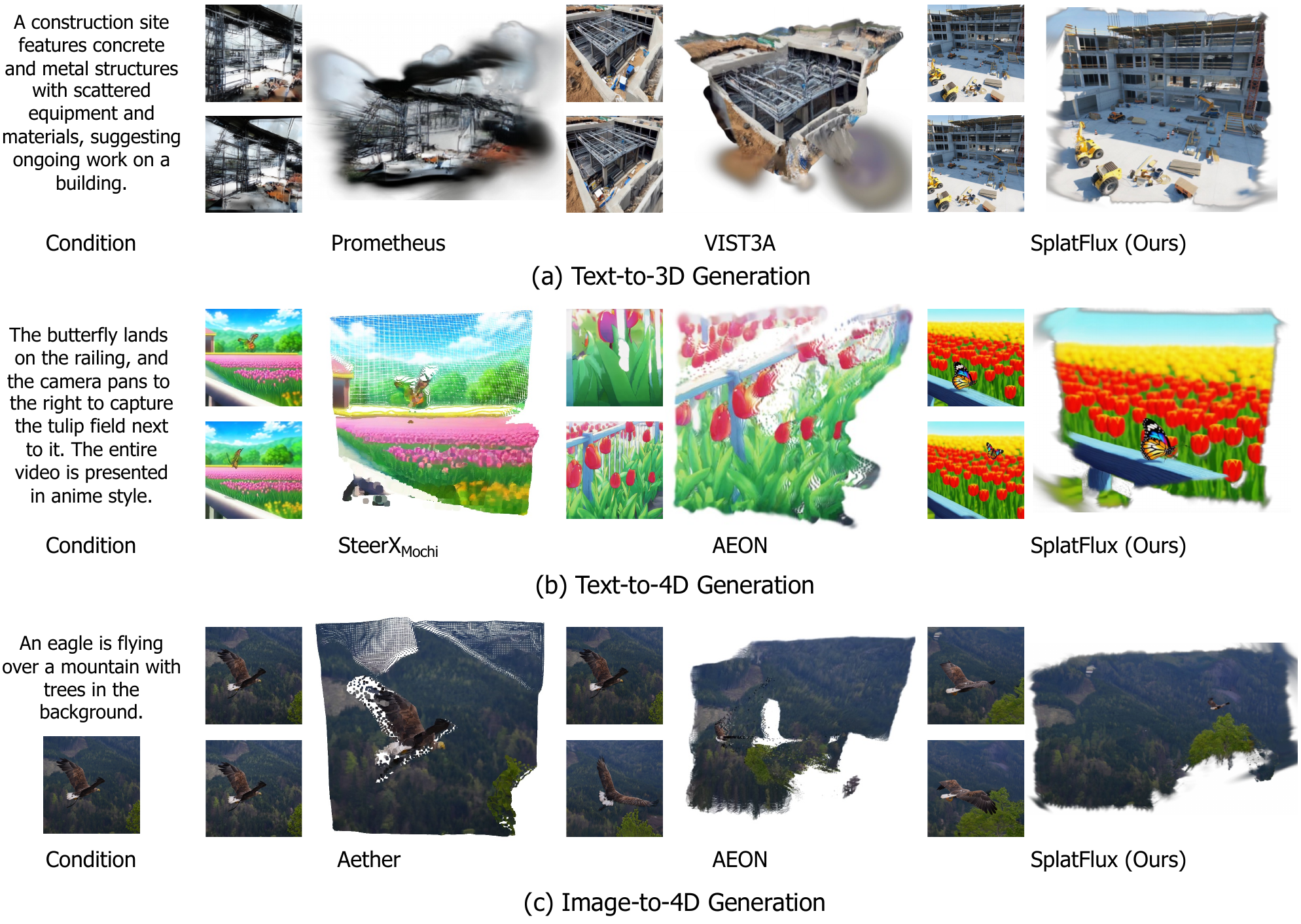} 
    \vspace{-2mm}
    \caption{Qualitative comparison of conditional 3D and 4D generation across different methods.}
    \label{fig:gen}
\end{figure*}

\subsection{Dynamic Object Segmentation}
\label{sec:dynamic_object_segmentation}
We further evaluate SplatFlux for dynamic object segmentation on the DAVIS~\citep{DAVIS} dataset. Quantitative metrics including mean IoU (JM), recall of IoU (JR), mean contour accuracy (FM), and recall of contour accuracy (FR) are reported in Table~\ref{tab:cmp_dyn_mask}. SplatFlux achieves the best performance, demonstrating its ability to distinguish dynamic regions. Qualitative results in Fig.~\ref{fig:dynmask} further show that SplatFlux produces more accurate and complete dynamic masks.

\subsection{Conditional 3D and 4D Generation}

\begin{table*}[!t]
    \centering
    \caption{\textbf{Quantitative comparison of text-to-3D generation} on SceneBench~\citep{Prometheus}. The \textbf{best} and \underline{second-best} results are highlighted.}
    \label{tab:cmp_text_to_3d}
    {
        \fontsize{9}{10.8}\selectfont   
        \setlength{\tabcolsep}{6.2pt}
        \renewcommand{\arraystretch}{1.1}
            \begin{tabular}{lcccccc}
            \toprule
            \textbf{Method} & Imaging $\uparrow$ & Aesthetic $\uparrow$ & CLIP $\uparrow$ & Alignment $\uparrow$ & Coherence $\uparrow$ & Style $\uparrow$  \\
            \midrule
            Prometheus~\citep{Prometheus} & 44.73 & 45.85 & 28.57 & 3.20 & 3.36 & 2.98 \\
            SplatFlow~\citep{SplatFlow} & 48.85 & 53.71 & 29.43 & 3.47 & 3.65 & 3.26 \\
            VIST3A~\citep{VIST3A} & \underline{64.87} & \underline{56.96} & \underline{30.18} & \underline{3.67} & \underline{3.86} & \underline{3.40} \\
            \midrule
            SplatFlux (Ours) & \textbf{69.18} & \textbf{58.08} & \textbf{31.35} & \textbf{3.72} & \textbf{3.97} & \textbf{3.48} \\
            \bottomrule
        \end{tabular}
    }
\end{table*}

\begin{table*}[!t]
    \centering
    \caption{\textbf{Quantitative comparison of text-to-4D generation} on PenguinVideoBenchmark~\citep{HunyuanVideo}. SteerX$_\textrm{Mochi/HunyuanVideo}$ denotes the variant using Mochi~\citep{Mochi} or HunyuanVideo~\citep{HunyuanVideo} as the base model. The \textbf{best} and \underline{second-best} results are highlighted for clarity.}
    \label{tab:cmp_text_to_4d}
    {
        \fontsize{9}{10.8}\selectfont   
        \setlength{\tabcolsep}{6.2pt}
        \renewcommand{\arraystretch}{1.1}
            \begin{tabular}{lccccc}
            \toprule
            \textbf{Method} & Sub. consistency $\uparrow$ & B.g. consistency $\uparrow$ & Dynamic degree $\uparrow$ & Aesthetic $\uparrow$ & Imaging $\uparrow$  \\
            \midrule
            SteerX$_\textrm{Mochi}$~\citep{SteerX} & 94.07 & 93.88 & 57.14 & 49.69 & 57.13\\
            SteerX$_\textrm{HunyuanVideo}$~\citep{SteerX} & \textbf{96.11} & \underline{95.23} & 37.76 & 53.32 & 61.23\\
            Wan2.1+NeoVerse~\citep{Wan, NeoVerse} & \underline{94.36} & 95.08 & \underline{69.39} & 53.71 & \underline{61.86} \\
            AEON~\citep{AEON} & 92.47 & 94.48 & 56.12 & \underline{53.79} & 56.90 \\
            \midrule
            SplatFlux (Ours) & \textbf{96.11} & \textbf{95.90} & \textbf{74.49} & \textbf{64.07} & \textbf{70.92}\\
            \bottomrule
        \end{tabular}
    }
\end{table*}

\begin{table*}[!t]
    \centering
    \caption{\textbf{Quantitative comparison of image-to-4D generation} on VBench-I2V~\citep{vbench}. SteerX$_\textrm{CogVideoX}$ denotes the variant using CogVideoX~\citep{CogVideoX} as the base model. The \textbf{best} and \underline{second-best} results are highlighted for clarity.}
    \label{tab:cmp_image_to_4d}
    {
        \fontsize{8.75}{10.5}\selectfont   
        \setlength{\tabcolsep}{6pt}
        \renewcommand{\arraystretch}{1.1}
            \begin{tabular}{lccccc}
            \toprule
            \textbf{Method} & Sub. consistency $\uparrow$ & B.g. consistency $\uparrow$ & Dynamic degree $\uparrow$ & Aesthetic $\uparrow$ & Imaging $\uparrow$  \\
            \midrule
            SteerX$_\textrm{CogVideoX}$~\citep{SteerX} & 95.91 & 95.43 & 31.55 & \underline{64.16} & \underline{71.55}\\
            Aether~\citep{Aether} & \textbf{97.97} & \textbf{97.03} & 19.72 & 63.89 & 69.85\\
            Wan2.1+NeoVerse~\citep{Wan, NeoVerse}	& 95.63 & 95.38 & \underline{46.48} & 61.46 & 71.22 \\
            AEON~\citep{AEON} & 95.04 & 95.36 & 43.38 & 60.77 & 65.49 \\
            \midrule
            SplatFlux (Ours) & \underline{95.95} & \underline{96.01} & \textbf{49.86} & \textbf{64.84} & \textbf{72.07}\\
            \bottomrule 
        \end{tabular}
    }
\end{table*}

\noindent\textbf{Text-to-3D Generation.} We evaluate SplatFlux for text-to-3D generation on SceneBench~\citep{Prometheus}. In alignment with VIST3A~\citep{VIST3A}, we employ imaging quality and aesthetic quality in VBench~\citep{vbench} for visual fidelity, CLIP Score~\citep{clipscore} for text alignment, and alignment, coherence, and style scores in UnifiedReward~\citep{UnifiedReward} for overall generation quality. As shown in Table~\ref{tab:cmp_text_to_3d}, SplatFlux consistently outperforms competing methods across all metrics. Qualitative results in Fig.~\ref{fig:gen}(a) further show that SplatFlux generates scenes that more faithfully match the input text conditions.

\noindent\textbf{Text-to-4D Generation.} Following SteerX~\citep{SteerX}, we evaluate text-to-4D generation on PenguinVideoBenchmark~\citep{HunyuanVideo} using VBench~\citep{vbench} metrics. Table~\ref{tab:cmp_text_to_4d} shows that SplatFlux achieves the best results across all metrics. Its consistent improvements over the Wan2.1+NeoVerse~\citep{Wan, NeoVerse} pipeline, which first generates videos and then reconstructs 4D Gaussian scenes, highlight the benefits of direct latent-to-4DGS decoding. Fig.~\ref{fig:gen}(b) further shows that the scenes generated by SplatFlux more faithfully follow the text conditions while maintaining coherent appearance and plausible motion.

\noindent\textbf{Image-to-4D Generation.} In alignment with SteerX~\citep{SteerX}, we evaluate image-to-4D generation on 355 samples from VBench-I2V~\citep{vbench}. SplatFlux achieves state-of-the-art visual quality and motion dynamics with competitive scene consistency, as reported in Table~\ref{tab:cmp_image_to_4d}. Fig.~\ref{fig:gen}(c) shows that SplatFlux generates more complete scenes with finer details than AEON~\citep{AEON} and more realistic and pronounced object motion than Aether~\citep{Aether}.

\subsection{Qualitative Visualization}
\label{sec:qualitative_visualization}
We visualize reconstructed and generated 4DGS scenes to examine their consistency under independent changes in camera pose and time.

\begin{figure*}[p]
    \centering
    \includegraphics[width=0.95\linewidth]{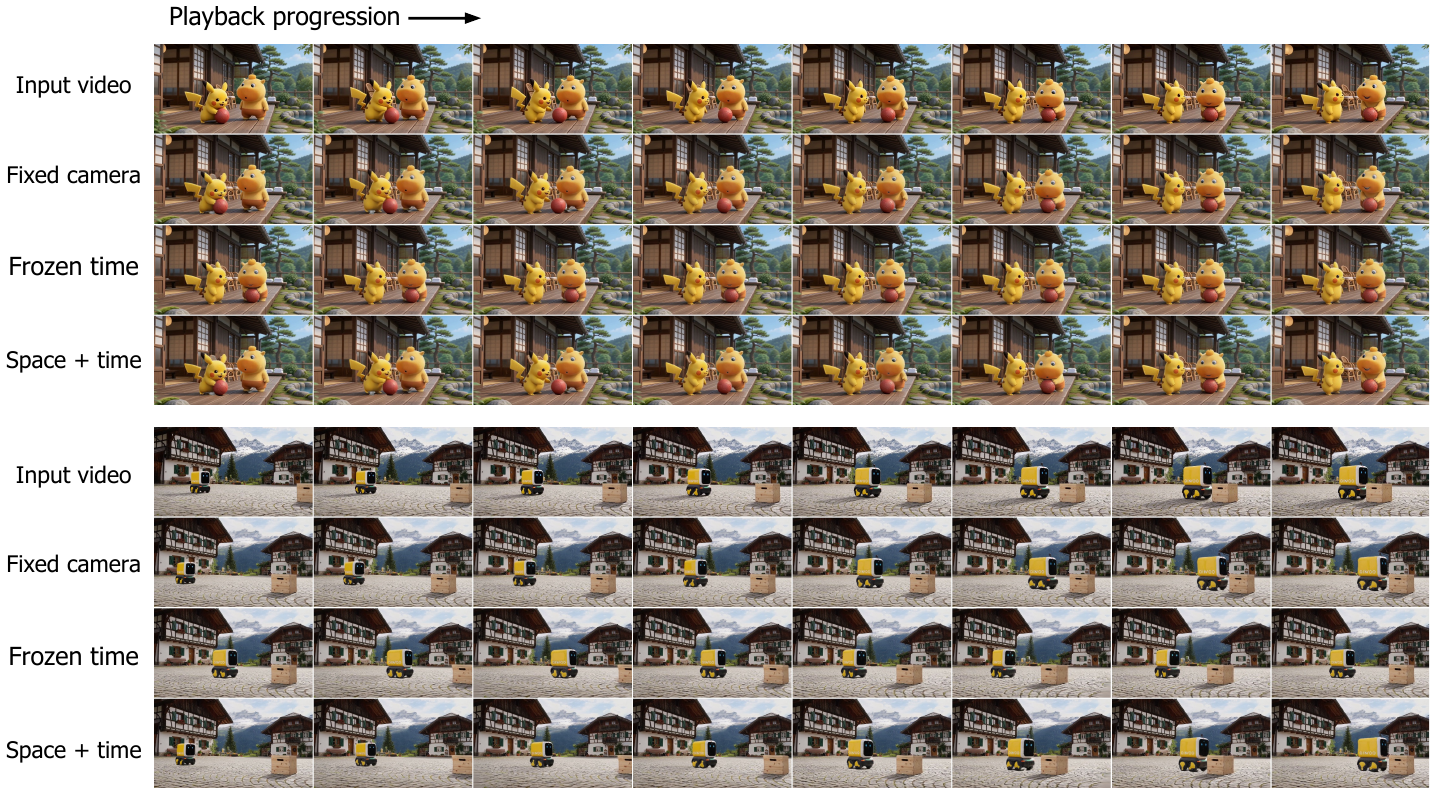} 
    \caption{Visualization of 4DGS scenes reconstructed by SplatFlux. Rows show the input video, fixed-camera rendering, frozen-time rendering, and joint camera–time variation. Columns follow the progression of each visualization sequence.}
    \label{fig:recon_demo}
\vspace{6pt}
    \includegraphics[width=0.95\linewidth]{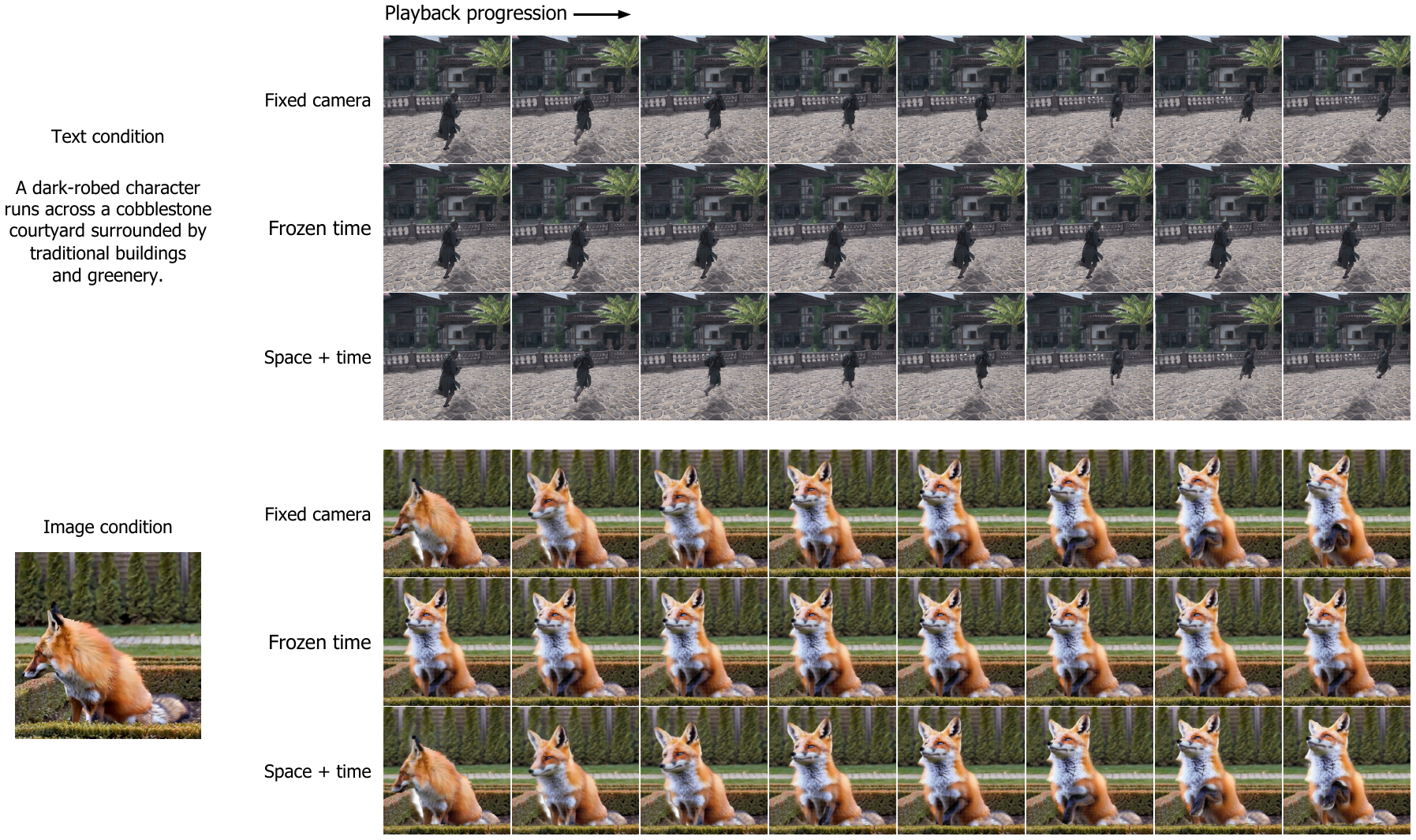}
    \caption{Visualization of 4DGS scenes generated by SplatFlux. The corresponding text or image condition is shown on the left. For each scene, rows show fixed-camera rendering, frozen-time rendering, and joint camera-time variation. Columns follow the progression of each visualization sequence.}
    \label{fig:gen_demo}
\end{figure*}

\noindent\textbf{Reconstructed 4DGS Scenes.}
Fig.~\ref{fig:recon_demo} visualizes scenes reconstructed by SplatFlux, demonstrating independent control of camera pose and time through fixed-camera, frozen-time, and joint camera-time rendering. The rendered sequences preserve scene appearance and motion, exhibiting visual consistency across changes in viewpoint and time.

\noindent\textbf{Generated 4DGS Scenes.}
Fig.~\ref{fig:gen_demo} visualizes scenes generated by SplatFlux under text and image conditions. The generated scenes reflect the conditioning content and maintain visual consistency across changes in viewpoint and time.

\subsection{Ablations}
\begin{table}[!t]
    \centering
        \caption{Ablations of static-dynamic decoupling and auxiliary supervision on the NVIDIA Dynamic Scene~\citep{NVIDIA_Dynamic_Scene} dataset.}
        \label{tab:ablation}
        \small
        \resizebox{\linewidth}{!}{
            \setlength{\tabcolsep}{4pt}
            \renewcommand{\arraystretch}{1.05}
            \begin{tabular}{lccc}
                \toprule
                \textbf{Settings} & PSNR $\uparrow$ & SSIM $\uparrow$ & LPIPS $\downarrow$ \\
                \midrule
                w/o static-dynamic decoupling & 13.94 & 0.498 & 0.466 \\
                w/o dynamic mask supervision & 18.41 & 0.530 & 0.266 \\
                w/o optical-flow supervision & 18.62 & 0.526 & 0.262 \\ 
                \midrule
                \textbf{Full Model} & \textbf{23.03} & \textbf{0.685} & \textbf{0.205} \\
                \bottomrule
            \end{tabular}
        }
\end{table}

\begin{figure}[!t]
    \centering
    \includegraphics[width=0.98\linewidth]{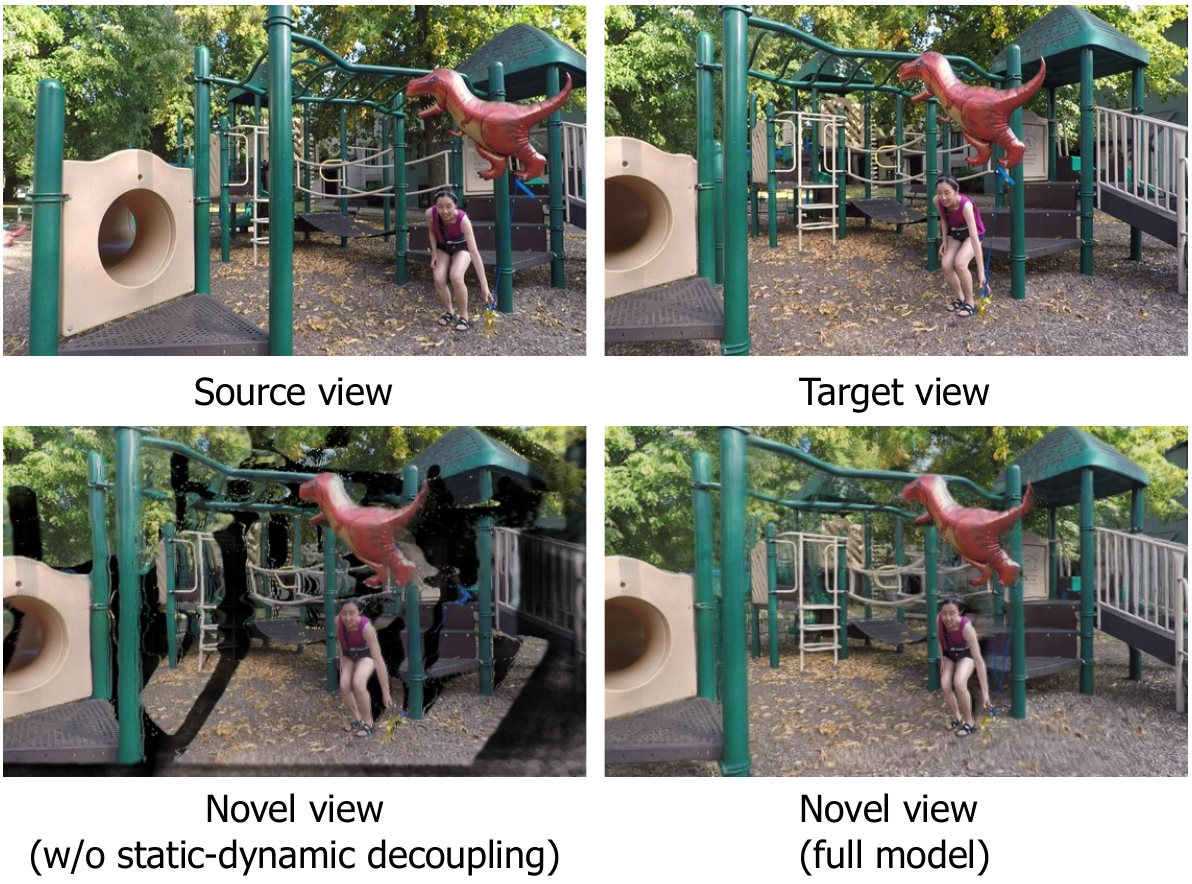}
    \caption{Qualitative results of the static-dynamic decoupling.}
    \label{fig:ablation}
\end{figure}

\begin{figure}[!t]
    \centering
    \includegraphics[width=0.98\linewidth]{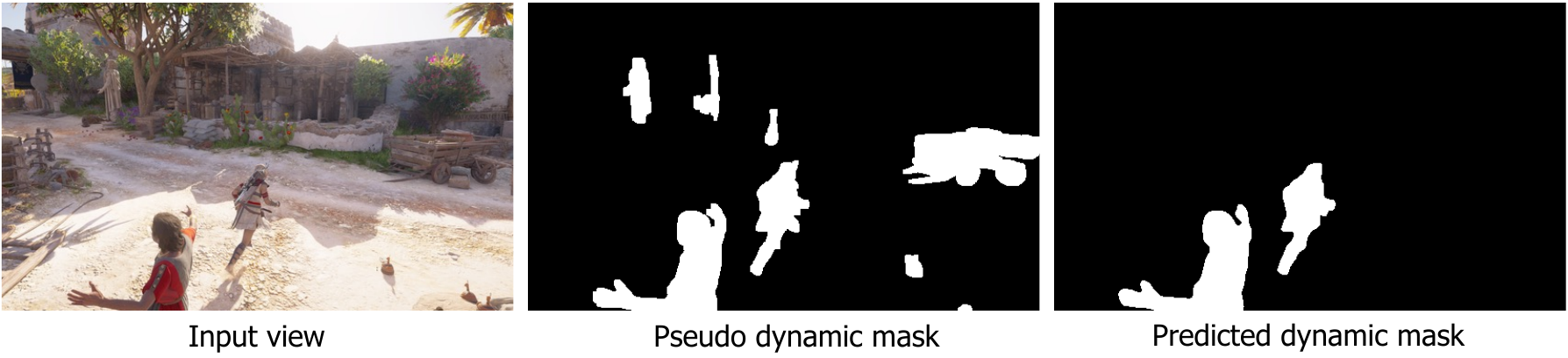}
    \caption{Qualitative comparison between the pseudo dynamic mask and the dynamic mask predicted by SplatFlux.}
    \label{fig:ablation_mask}
\end{figure}

We present ablation studies on our static-dynamic decoupled representation and auxiliary supervision using the NVIDIA Dynamic Scene dataset in Table~\ref{tab:ablation}. Removing the decoupled representation causes the largest performance drop, suggesting that explicit static-dynamic separation is essential for complete 4D scene construction. Without this constraint, the model tends to learn an interpolation-like representation from monocular videos, where static structures may be treated as dynamic, leading to incomplete geometry and holes under novel view rendering, as illustrated in Fig.~\ref{fig:ablation}. The two supervision signals further provide complementary constraints for dynamic attributes. Dynamic mask supervision guides where the scene should be decomposed, while optical-flow supervision constrains how dynamic Gaussians should move. Removing either supervision signal degrades performance.

Notably, SplatFlux does not require perfectly accurate pseudo labels, benefiting from the consistency constraints provided by rendering supervision. Fig.~\ref{fig:ablation_mask} presents a qualitative comparison between the pseudo dynamic mask used for supervision and the dynamic mask predicted by SplatFlux. Compared with the pseudo mask, SplatFlux produces cleaner dynamic regions and avoids incorrectly classifying static scene structures as dynamic, indicating that the learned static-dynamic decomposition is more robust than the noisy supervision.

\section{Limitations}
SplatFlux currently works in an offline setting and does not support streaming reconstruction or generation. We leave streaming latent decoding to future work.

\section{Conclusions}
We have presented SplatFlux, a unified feed-forward latent-to-4DGS framework for video-based 4D reconstruction and conditional 4D generation. By decoding video latents from different sources into a 4D Gaussian representation, SplatFlux provides a shared formulation for captured and generated dynamic content. We further introduce a static-dynamic decoupled 4D Gaussian representation that supports complete 4D scene construction in a feed-forward manner. Through auxiliary distillation from vision-language-guided dynamic masks and optical-flow cues, SplatFlux learns dynamic attributes without costly annotations. Extensive experiments on 3D and 4D reconstruction and generation demonstrate the effectiveness and generality of SplatFlux.

\section*{Acknowledgments}
This work was supported by Ant Group Research Intern Program.

\clearpage

{
    \small
    \bibliographystyle{ieeenat_fullname}
    \bibliography{main}

@inproceedings{SfM,
  title={Structure-from-motion revisited},
  author={Sch{\"o}nberger, Johannes L and Frahm, Jan-Michael},
  booktitle={Proceedings of the IEEE Conference on Computer Vision and Pattern Recognition},
  pages={4104--4113},
  year={2016}
}

@inproceedings{NeRF,
  title={{NeRF}: Representing scenes as neural radiance fields for view synthesis},
  author={Ben Mildenhall and Pratul P. Srinivasan and Matthew Tancik and Jonathan T. Barron and Ravi Ramamoorthi and Ren Ng},
  booktitle={European Conference on Computer Vision},
  year={2020},
}

@article{3DGS,
  title={{3D} {Gaussian} Splatting for Real-Time Radiance Field Rendering},
  author={Kerbl, Bernhard and Kopanas, Georgios and Leimk{\"u}hler, Thomas and Drettakis, George},
  journal={ACM Transactions on Graphics},
  volume={42},
  number={4},
  pages={139:1--139:14},
  year={2023}
}

@inproceedings{Shape_of_motion,
  title={{Shape of Motion}: {4D} reconstruction from a single video},
  author={Wang, Qianqian and Ye, Vickie and Gao, Hang and Zeng, Weijia and Austin, Jake and Li, Zhengqi and Kanazawa, Angjoo},
  booktitle={Proceedings of the IEEE/CVF International Conference on Computer Vision},
  pages={9660--9672},
  year={2025}
}

@inproceedings{MoSca,
  title={{MoSca}: Dynamic {Gaussian} fusion from casual videos via {4D} motion scaffolds},
  author={Lei, Jiahui and Weng, Yijia and Harley, Adam W and Guibas, Leonidas and Daniilidis, Kostas},
  booktitle={Proceedings of the IEEE/CVF Conference on Computer Vision and Pattern Recognition},
  pages={6165--6177},
  year={2025}
}

@article{Splatter_a_video,
  title={Splatter a video: Video {Gaussian} representation for versatile processing},
  author={Sun, Yang-Tian and Huang, Yi-Hua and Ma, Lin and Lyu, Xiaoyang and Cao, Yan-Pei and Qi, Xiaojuan},
  journal={Advances in Neural Information Processing Systems},
  volume={37},
  pages={50401--50425},
  year={2024}
}

@inproceedings{RetimeGS,
  title={{RetimeGS}: Continuous-Time Reconstruction of {4D} {Gaussian} Splatting},
  author={Wang, Xuezhen and Ma, Li and Shen, Yulin and Wang, Zeyu and Sander, Pedro V},
  year={2026},
  booktitle={Proceedings of the IEEE/CVF Conference on Computer Vision and Pattern Recognition}
}

@inproceedings{ViT,
  title={An image is worth 16x16 words: Transformers for image recognition at scale},
  author={Dosovitskiy, Alexey and Beyer, Lucas and Kolesnikov, Alexander and Weissenborn, Dirk and Zhai, Xiaohua and Unterthiner, Thomas and Dehghani, Mostafa and Minderer, Matthias and Heigold, Georg and Gelly, Sylvain and others},
  year={2021},
  booktitle={International Conference on Learning Representations}
}

@article{DINOv2,
  title={{DINOv2}: Learning robust visual features without supervision},
  author={Oquab, Maxime and Darcet, Timoth{\'e}e and Moutakanni, Th{\'e}o and Vo, Huy and Szafraniec, Marc and Khalidov, Vasil and Fernandez, Pierre and Haziza, Daniel and Massa, Francisco and El-Nouby, Alaaeldin and others},
  journal={Transactions on Machine Learning Research},
  year={2024}
}

@article{DINOv3,
  title={{DINOv3}},
  author={Sim{\'e}oni, Oriane and Vo, Huy V and Seitzer, Maximilian and Baldassarre, Federico and Oquab, Maxime and Jose, Cijo and Khalidov, Vasil and Szafraniec, Marc and Yi, Seungeun and Ramamonjisoa, Micha{\"e}l and others},
  journal={arXiv preprint arXiv:2508.10104},
  year={2025}
}

@inproceedings{DPT,
  title={Vision transformers for dense prediction},
  author={Ranftl, Ren{\'e} and Bochkovskiy, Alexey and Koltun, Vladlen},
  booktitle={Proceedings of the IEEE/CVF International Conference on Computer Vision},
  pages={12179--12188},
  year={2021}
}

@inproceedings{DUSt3R,
  author={Shuzhe Wang and Vincent Leroy and Yohann Cabon and Boris Chidlovskii and J{\'{e}}r{\^{o}}me Revaud},
  title={{DUSt3R}: Geometric {3D} Vision Made Easy},
  booktitle={{Proceedings of the IEEE/CVF Conference on Computer Vision and Pattern Recognition}},
  pages={20697--20709},
  year={2024}
}

@inproceedings{MASt3R,
  title={Grounding image matching in {3D} with {MASt3R}},
  author={Leroy, Vincent and Cabon, Yohann and Revaud, J{\'e}r{\^o}me},
  booktitle={European Conference on Computer Vision},
  pages={71--91},
  year={2024}
}

@InProceedings{MoGe,
    author={Wang, Ruicheng and Xu, Sicheng and Dai, Cassie and Xiang, Jianfeng and Deng, Yu and Tong, Xin and Yang, Jiaolong},
    title={{MoGe}: Unlocking Accurate Monocular Geometry Estimation for Open-Domain Images with Optimal Training Supervision},
    booktitle={Proceedings of the IEEE/CVF Conference on Computer Vision and Pattern Recognition},
    year={2025},
    pages={5261--5271}
}

@inproceedings{DA3,
    title={{Depth Anything 3}: Recovering the Visual Space from Any Views},
    author={Haotong Lin and Sili Chen and Jun Hao Liew and Donny Y. Chen and Zhenyu Li and Yang Zhao and Sida Peng and Hengkai Guo and Xiaowei Zhou and Guang Shi and Jiashi Feng and Bingyi Kang},
    booktitle={International Conference on Learning Representations},
    year={2026},
}

@article{Dens3R,
    title={{Dens3R}: A Foundation Model for {3D} Geometry Prediction}, 
    author={Xianze Fang and Jingnan Gao and Zhe Wang and Zhuo Chen and Xingyu Ren and Jiangjing Lyu and Qiaomu Ren and Zhonglei Yang and Xiaokang Yang and Yichao Yan and Chengfei Lyu},
    journal={arXiv preprint arXiv:2507.16290},
    year={2025}
}

@inproceedings{MoRE,
  title={{MoRE}: {3D} Visual Geometry Reconstruction Meets Mixture-of-Experts}, 
  author={Jingnan Gao and Zhe Wang and Xianze Fang and Xingyu Ren and Zhuo Chen and Shengqi Liu and Yuhao Cheng and Jiangjing Lyu and Xiaokang Yang and Yichao Yan},
  year={2026},
  booktitle={Proceedings of the IEEE/CVF Conference on Computer Vision and Pattern Recognition},
  pages={14680--14691}
}

@inproceedings{LDCM,
    title={Large Depth Completion Model from Sparse Observations},
    author={Zhu Yu and Zhengyi Zhao and Runmin Zhang and Lingteng Qiu and Si-Yuan Cao and Kejie Qiu and Yisheng He and Siyu Zhu and Zilong Dong and Hui-Liang Shen},
    booktitle={International Conference on Learning Representations},
    year={2026}
}

@inproceedings{VGGT,
  title={{VGGT}: Visual geometry grounded transformer},
  author={Wang, Jianyuan and Chen, Minghao and Karaev, Nikita and Vedaldi, Andrea and Rupprecht, Christian and Novotny, David},
  booktitle={Proceedings of the IEEE/CVF Conference on Computer Vision and Pattern Recognition},
  pages={5294--5306},
  year={2025}
}

@inproceedings{Pi3,
title={{$\pi^3$}: Permutation-Equivariant Visual Geometry Learning},
author={Yifan Wang and Jianjun Zhou and Haoyi Zhu and Wenzheng Chang and Yang Zhou and Zizun Li and Junyi Chen and Jiangmiao Pang and Chunhua Shen and Tong He},
booktitle={International Conference on Learning Representations},
year={2026}
}

@inproceedings{MonST3R,
  title={{MonST3R}: A Simple Approach for Estimating Geometry in the Presence of Motion},
  author={Zhang, Junyi and Herrmann, Charles and Hur, Junhwa and Jampani, Varun and Darrell, Trevor and Cole, Forrester and Sun, Deqing and Yang, Ming-Hsuan},
  booktitle={International Conference on Learning Representations},
  year={2025},
}

@inproceedings{FLARE,
  title={{FLARE}: Feed-forward Geometry, Appearance and Camera Estimation from Uncalibrated Sparse Views}, 
  author={Shangzhan Zhang and Jianyuan Wang and Yinghao Xu and Nan Xue and Christian Rupprecht and Xiaowei Zhou and Yujun Shen and Gordon Wetzstein},
  year = {2025},
  booktitle={Proceedings of the IEEE/CVF Conference on Computer Vision and Pattern Recognition},
  pages={21936--21947}
}

@inproceedings{CUT3R,
  title={Continuous {3D} perception model with persistent state},
  author={Wang, Qianqian and Zhang, Yifei and Holynski, Aleksander and Efros, Alexei A and Kanazawa, Angjoo},
  booktitle={Proceedings of the IEEE/CVF Conference on Computer Vision and Pattern Recognition},
  pages={10510--10522},
  year={2025}
}

@inproceedings{Easi3R,
  title={{Easi3R}: Estimating disentangled motion from {DUSt3R} without training},
  author={Chen, Xingyu and Chen, Yue and Xiu, Yuliang and Geiger, Andreas and Chen, Anpei},
  booktitle={Proceedings of the IEEE/CVF International Conference on Computer Vision},
  pages={9158--9168},
  year={2025}
}

@article{4D-VGGT,
  title={{4D-VGGT}: A General Foundation Model with SpatioTemporal Awareness for Dynamic Scene Geometry Estimation},
  author={Wang, Haonan and Zhou, Hanyu and Liu, Haoyue and Yan, Luxin},
  journal={arXiv preprint arXiv:2511.18416},
  year={2025}
}

@article{VGGT4D,
  title={{VGGT4D}: Mining Motion Cues in Visual Geometry Transformers for {4D} Scene Reconstruction},
  author={Hu, Yu and Cheng, Chong and Yu, Sicheng and Guo, Xiaoyang and Wang, Hao},
  journal={arXiv preprint arXiv:2511.19971},
  year={2025}
}

@inproceedings{VDA,
  title={{Video Depth Anything}: Consistent depth estimation for super-long videos},
  author={Chen, Sili and Guo, Hengkai and Zhu, Shengnan and Zhang, Feihu and Huang, Zilong and Feng, Jiashi and Kang, Bingyi},
  booktitle={Proceedings of the IEEE/CVF Conference on Computer Vision and Pattern Recognition},
  pages={22831--22840},
  year={2025}
}

@inproceedings{DepthCrafter,
  title={{DepthCrafter}: Generating consistent long depth sequences for open-world videos},
  author={Hu, Wenbo and Gao, Xiangjun and Li, Xiaoyu and Zhao, Sijie and Cun, Xiaodong and Zhang, Yong and Quan, Long and Shan, Ying},
  booktitle={Proceedings of the IEEE/CVF Conference on Computer Vision and Pattern Recognition},
  pages={2005--2015},
  year={2025}
}

@inproceedings{GeometryCrafter,
  title={{GeometryCrafter}: Consistent geometry estimation for open-world videos with diffusion priors},
  author={Xu, Tian-Xing and Gao, Xiangjun and Hu, Wenbo and Li, Xiaoyu and Zhang, Song-Hai and Shan, Ying},
  booktitle={Proceedings of the IEEE/CVF International Conference on Computer Vision},
  pages={6632--6644},
  year={2025}
}

@article{AnySplat,
  title={{AnySplat}: Feed-forward {3D} {Gaussian} splatting from unconstrained views},
  author={Jiang, Lihan and Mao, Yucheng and Xu, Linning and Lu, Tao and Ren, Kerui and Jin, Yichen and Xu, Xudong and Yu, Mulin and Pang, Jiangmiao and Zhao, Feng and Lin, Dahua and Dai, Bo},
  journal={ACM Transactions on Graphics},
  volume={44},
  number={6},
  pages={257:1--257:16},
  year={2025},
}

@article{DynamicVerse,
  title={{DynamicVerse}: A Physically-Aware Multimodal Framework for {4D} World Modeling},
  author={Wen, Kairun and Huang, Yuzhi and Chen, Runyu and Zheng, Hui and Lin, Yunlong and Pan, Panwang and Li, Chenxin and Cong, Wenyan and Zhang, Jian and Lu, Junbin and others},
  journal={arXiv preprint arXiv:2512.03000},
  year={2025}
}

@inproceedings{WorldMirror,
  title={{WorldMirror}: Universal {3D} world reconstruction with any-prior prompting},
  author={Liu, Yifan and Min, Zhiyuan and Wang, Zhenwei and Wu, Junta and Wang, Tengfei and Yuan, Yixuan and Luo, Yawei and Guo, Chunchao},
  year={2026},
  booktitle={International Conference on Machine Learning}
}

@inproceedings{NoPoSplat,
  title={No pose, no problem: Surprisingly simple {3D} {Gaussian} splats from sparse unposed images},
  author={Ye, Botao and Liu, Sifei and Xu, Haofei and Li, Xueting and Pollefeys, Marc and Yang, Ming-Hsuan and Peng, Songyou},
  year={2025},
  booktitle={International Conference on Learning Representations}
}

@article{YoNoSplat,
  title={{YoNoSplat}: You Only Need One Model for Feedforward {3D} {Gaussian} Splatting},
  author={Ye, Botao and Chen, Boqi and Xu, Haofei and Barath, Daniel and Pollefeys, Marc},
  journal={arXiv preprint arXiv:2511.07321},
  year={2025}
}

@inproceedings{STORM,
  title   = {{STORM}: Spatio-Temporal Reconstruction Model for Large-Scale Outdoor Scenes},
  author  = {Jiawei Yang and Jiahui Huang and Yuxiao Chen and Yan Wang and Boyi Li and Yurong You and Maximilian Igl and Apoorva Sharma and Peter Karkus and Danfei Xu and Boris Ivanovic and Yue Wang and Marco Pavone},
  booktitle={International Conference on Learning Representations},
  year    = {2025}
}

@article{4DGT,
  title={{4DGT}: Learning a {4D} {Gaussian} transformer using real-world monocular videos},
  author={Xu, Zhen and Li, Zhengqin and Dong, Zhao and Zhou, Xiaowei and Newcombe, Richard and Lv, Zhaoyang},
  journal={arXiv preprint arXiv:2506.08015},
  year={2025}
}

@inproceedings{MoVieS,
  title={{MoVieS}: Motion-aware {4D} dynamic view synthesis in one second},
  author={Lin, Chenguo and Lin, Yuchen and Pan, Panwang and Yu, Yifan and Hu, Tao and Yan, Honglei and Fragkiadaki, Katerina and Mu, Yadong},
  year={2026},
  booktitle={Proceedings of the IEEE/CVF Conference on Computer Vision and Pattern Recognition}
}

@inproceedings{NeoVerse,
  title={{NeoVerse}: Enhancing {4D} World Model with in-the-wild Monocular Videos},
  author={Yang, Yuxue and Fan, Lue and Shi, Ziqi and Peng, Junran and Wang, Feng and Zhang, Zhaoxiang},
  year={2026},
  booktitle={Proceedings of the IEEE/CVF Conference on Computer Vision and Pattern Recognition}
}

@inproceedings{L4GM,
  title={{L4GM}: Large {4D} {Gaussian} Reconstruction Model},
  author={Ren, Jiawei and Xie, Kevin and Mirzaei, Ashkan and Liang, Hanxue and Zeng, Xiaohui and Kreis, Karsten and Liu, Ziwei and Torralba, Antonio and Fidler, Sanja and Kim, Seung Wook and Ling, Huan},
  booktitle={Advances in Neural Information Processing Systems},
  year={2024}
}

@article{DGS-LRM,
  title={{DGS-LRM}: Real-time deformable {3D} {Gaussian} reconstruction from monocular videos},
  author={Lin, Chieh Hubert and Lv, Zhaoyang and Wu, Songyin and Xu, Zhen and Nguyen-Phuoc, Thu and Tseng, Hung-Yu and Straub, Julian and Khan, Numair and Xiao, Lei and Yang, Ming-Hsuan and others},
  journal={arXiv preprint arXiv:2506.09997},
  year={2025}
}

@article{UFO-4D,
  title={{UFO-4D}: Unposed Feedforward {4D} Reconstruction from Two Images},
  author={Hur, Junhwa and Herrmann, Charles and Peng, Songyou and Henzler, Philipp and Ma, Zeyu and Zickler, Todd and Sun, Deqing},
  journal={arXiv preprint arXiv:2602.24290},
  year={2026}
}

@article{BTimer,
  title={Feed-forward bullet-time reconstruction of dynamic scenes from monocular videos},
  author={Liang, Hanxue and Ren, Jiawei and Mirzaei, Ashkan and Torralba, Antonio and Liu, Ziwei and Gilitschenski, Igor and Fidler, Sanja and Oztireli, Cengiz and Ling, Huan and Gojcic, Zan and others},
  journal={arXiv preprint arXiv:2412.03526},
  year={2024}
}

@article{StreamSplat,
  title={{StreamSplat}: Towards Online Dynamic {3D} Reconstruction from Uncalibrated Video Streams},
  author={Wu, Zike and Yan, Qi and Yi, Xuanyu and Wang, Lele and Liao, Renjie},
  journal={arXiv preprint arXiv:2506.08862},
  year={2025}
}

@inproceedings{StableDiffusion,
  title={High-resolution image synthesis with latent diffusion models},
  author={Rombach, Robin and Blattmann, Andreas and Lorenz, Dominik and Esser, Patrick and Ommer, Bj{\"o}rn},
  booktitle={Proceedings of the IEEE/CVF Conference on Computer Vision and Pattern Recognition},
  pages={10684--10695},
  year={2022}
}

@article{Imagen,
  title={Photorealistic text-to-image diffusion models with deep language understanding},
  author={Saharia, Chitwan and Chan, William and Saxena, Saurabh and Li, Lala and Whang, Jay and Denton, Emily L and Ghasemipour, Kamyar and Gontijo Lopes, Raphael and Karagol Ayan, Burcu and Salimans, Tim and others},
  journal={Advances in Neural Information Processing Systems},
  volume={35},
  pages={36479--36494},
  year={2022}
}

@article{Wan,
  title={{Wan}: Open and advanced large-scale video generative models},
  author={{Team Wan} and Wang, Ang and Ai, Baole and Wen, Bin and Mao, Chaojie and Xie, Chen-Wei and Chen, Di and Yu, Feiwu and Zhao, Haiming and Yang, Jianxiao and others},
  journal={arXiv preprint arXiv:2503.20314},
  year={2025}
}

@inproceedings{CogVideoX,
  title={{CogVideoX}: Text-to-video diffusion models with an expert transformer},
  author={Yang, Zhuoyi and Teng, Jiayan and Zheng, Wendi and Ding, Ming and Huang, Shiyu and Xu, Jiazheng and Yang, Yuanming and Hong, Wenyi and Zhang, Xiaohan and Feng, Guanyu and others},
  year={2025},
  booktitle={International Conference on Learning Representations}
}

@article{DreamFusion,
  title={{DreamFusion}: Text-to-{3D} using {2D} diffusion},
  author={Poole, Ben and Jain, Ajay and Barron, Jonathan T and Mildenhall, Ben},
  journal={arXiv preprint arXiv:2209.14988},
  year={2022}
}

@article{ProlificDreamer,
  title={{ProlificDreamer}: High-fidelity and diverse text-to-{3D} generation with variational score distillation},
  author={Wang, Zhengyi and Lu, Cheng and Wang, Yikai and Bao, Fan and Li, Chongxuan and Su, Hang and Zhu, Jun},
  journal={Advances in Neural Information Processing Systems},
  volume={36},
  pages={8406--8441},
  year={2023}
}

@inproceedings{MAV3D,
  title={Text-to-{4D} dynamic scene generation},
  author={Singer, Uriel and Sheynin, Shelly and Polyak, Adam and Ashual, Oron and Makarov, Iurii and Kokkinos, Filippos and Goyal, Naman and Vedaldi, Andrea and Parikh, Devi and Johnson, Justin and others},
  year={2023},
  booktitle={Proceedings of the 40th International Conference on Machine Learning},
  volume={202},
  pages={31915--31929}
}

@inproceedings{TC4D,
  title={{TC4D}: Trajectory-conditioned text-to-{4D} generation},
  author={Bahmani, Sherwin and Liu, Xian and Yifan, Wang and Skorokhodov, Ivan and Rong, Victor and Liu, Ziwei and Liu, Xihui and Park, Jeong Joon and Tulyakov, Sergey and Wetzstein, Gordon and others},
  booktitle={European Conference on Computer Vision},
  pages={53--72},
  year={2024},
}

@inproceedings{4D-fy,
  title={{4D-fy}: Text-to-{4D} generation using hybrid score distillation sampling},
  author={Bahmani, Sherwin and Skorokhodov, Ivan and Rong, Victor and Wetzstein, Gordon and Guibas, Leonidas and Wonka, Peter and Tulyakov, Sergey and Park, Jeong Joon and Tagliasacchi, Andrea and Lindell, David B},
  booktitle={Proceedings of the IEEE/CVF Conference on Computer Vision and Pattern Recognition},
  pages={7996--8006},
  year={2024}
}

@inproceedings{Bolt3D,
  title={{Bolt3D}: Generating {3D} scenes in seconds},
  author={Szymanowicz, Stanislaw and Zhang, Jason Y and Srinivasan, Pratul and Gao, Ruiqi and Brussee, Arthur and Holynski, Aleksander and Martin-Brualla, Ricardo and Barron, Jonathan T and Henzler, Philipp},
  booktitle={Proceedings of the IEEE/CVF International Conference on Computer Vision},
  pages={24846--24857},
  year={2025}
}

@inproceedings{LGM,
  title={{LGM}: Large multi-view {Gaussian} model for high-resolution {3D} content creation},
  author={Tang, Jiaxiang and Chen, Zhaoxi and Chen, Xiaokang and Wang, Tengfei and Zeng, Gang and Liu, Ziwei},
  booktitle={European Conference on Computer Vision},
  pages={1--18},
  year={2024},
  organization={Springer}
}

@article{CAT3D,
  title={{CAT3D}: Create anything in {3D} with multi-view diffusion models},
  author={Gao, Ruiqi and Holynski, Aleksander and Henzler, Philipp and Brussee, Arthur and Martin-Brualla, Ricardo and Srinivasan, Pratul and Barron, Jonathan T and Poole, Ben},
  journal={Advances in Neural Information Processing Systems},
  year={2024},
  volume={37}
}

@article{DimensionX,
  title={{DimensionX}: Create any {3D} and {4D} scenes from a single image with controllable video diffusion},
  author={Sun, Wenqiang and Chen, Shuo and Liu, Fangfu and Chen, Zilong and Duan, Yueqi and Zhang, Jun and Wang, Yikai},
  journal={arXiv preprint arXiv:2411.04928},
  year={2024}
}

@inproceedings{SplatFlow,
  title={{SplatFlow}: Multi-view rectified flow model for {3D} {Gaussian} splatting synthesis},
  author={Go, Hyojun and Park, Byeongjun and Jang, Jiho and Kim, Jin-Young and Kwon, Soonwoo and Kim, Changick},
  booktitle={Proceedings of the IEEE/CVF Conference on Computer Vision and Pattern Recognition},
  pages={21524--21536},
  year={2025}
}

@inproceedings{SEVA,
  title={Stable virtual camera: Generative view synthesis with diffusion models},
  author={Zhou, Jensen and Gao, Hang and Voleti, Vikram and Vasishta, Aaryaman and Yao, Chun-Han and Boss, Mark and Torr, Philip and Rupprecht, Christian and Jampani, Varun},
  booktitle={Proceedings of the IEEE/CVF International Conference on Computer Vision},
  pages={12405--12414},
  year={2025}
}

@inproceedings{Prometheus,
  title={{Prometheus}: {3D}-aware latent diffusion models for feed-forward text-to-{3D} scene generation},
  author={Yang, Yuanbo and Shao, Jiahao and Li, Xinyang and Shen, Yujun and Geiger, Andreas and Liao, Yiyi},
  booktitle={Proceedings of the IEEE/CVF Conference on Computer Vision and Pattern Recognition},
  pages={2857--2869},
  year={2025}
}

@inproceedings{DiffusionGS,
  title={Baking {Gaussian} splatting into diffusion denoiser for fast and scalable single-stage image-to-{3D} generation and reconstruction},
  author={Cai, Yuanhao and Zhang, He and Zhang, Kai and Liang, Yixun and Ren, Mengwei and Luan, Fujun and Liu, Qing and Kim, Soo Ye and Zhang, Jianming and Zhang, Zhifei and others},
  booktitle={Proceedings of the IEEE/CVF International Conference on Computer Vision},
  pages={25062--25072},
  year={2025}
}

@inproceedings{VIST3A,
  title={Text-to-{3D} by Stitching a Multi-view Reconstruction Network to a Video Generator},
  author={Go, Hyojun and Narnhofer, Dominik and Bhat, Goutam and Truong, Prune and Tombari, Federico and Schindler, Konrad},
  year={2026},
  booktitle={International Conference on Learning Representations}
}

@inproceedings{CAT4D,
  title={{CAT4D}: Create anything in {4D} with multi-view video diffusion models},
  author={Wu, Rundi and Gao, Ruiqi and Poole, Ben and Trevithick, Alex and Zheng, Changxi and Barron, Jonathan T and Holynski, Aleksander},
  booktitle={Proceedings of the IEEE/CVF Conference on Computer Vision and Pattern Recognition},
  pages={26057--26068},
  year={2025}
}

@inproceedings{SV4D,
  title={{SV4D}: Dynamic {3D} content generation with multi-frame and multi-view consistency},
  author={Xie, Yiming and Yao, Chun-Han and Voleti, Vikram and Jiang, Huaizu and Jampani, Varun},
  year={2025},
  booktitle={International Conference on Learning Representations}
}

@inproceedings{4Real-Video,
  title={{4Real-Video}: Learning generalizable photo-realistic {4D} video diffusion},
  author={Wang, Chaoyang and Zhuang, Peiye and Ngo, Tuan Duc and Menapace, Willi and Siarohin, Aliaksandr and Vasilkovsky, Michael and Skorokhodov, Ivan and Tulyakov, Sergey and Wonka, Peter and Lee, Hsin-Ying},
  booktitle={Proceedings of the IEEE/CVF Conference on Computer Vision and Pattern Recognition},
  pages={17723--17732},
  year={2025}
}

@article{4DNeX,
  title={{4DNeX}: Feed-forward {4D} generative modeling made easy},
  author={Chen, Zhaoxi and Liu, Tianqi and Zhuo, Long and Ren, Jiawei and Tao, Zeng and Zhu, He and Hong, Fangzhou and Pan, Liang and Liu, Ziwei},
  journal={arXiv preprint arXiv:2508.13154},
  year={2025}
}

@inproceedings{SteerX,
  title={{SteerX}: Creating any camera-free {3D} and {4D} scenes with geometric steering},
  author={Park, Byeongjun and Go, Hyojun and Nam, Hyelin and Kim, Byung-Hoon and Chung, Hyungjin and Kim, Changick},
  booktitle={Proceedings of the IEEE/CVF International Conference on Computer Vision},
  pages={27326--27337},
  year={2025}
}

@inproceedings{Aether,
  title={{Aether}: Geometric-aware unified world modeling},
  author={Zhu, Haoyi and Wang, Yifan and Zhou, Jianjun and Chang, Wenzheng and Zhou, Yang and Li, Zizun and Chen, Junyi and Shen, Chunhua and Pang, Jiangmiao and He, Tong},
  booktitle={Proceedings of the IEEE/CVF International Conference on Computer Vision},
  pages={8535--8546},
  year={2025}
}

@article{Diff4Splat,
  title={{Diff4Splat}: Controllable {4D} Scene Generation with Latent Dynamic Reconstruction Models},
  author={Pan, Panwang and Lin, Chenguo and Zhao, Jingjing and Li, Chenxin and Lin, Yuchen and Li, Haopeng and Yan, Honglei and Wen, Kairun and Lin, Yunlong and Yuan, Yixuan and others},
  journal={arXiv preprint arXiv:2511.00503},
  year={2025}
}

@inproceedings{Lyra,
  title={{Lyra}: Generative {3D} Scene Reconstruction via Video Diffusion Model Self-Distillation},
  author={Bahmani, Sherwin and Shen, Tianchang and Ren, Jiawei and Huang, Jiahui and Jiang, Yifeng and Turki, Haithem and Tagliasacchi, Andrea and Lindell, David B. and Gojcic, Zan and Fidler, Sanja and Ling, Huan and Gao, Jun and Ren, Xuanchi},
  booktitle={International Conference on Learning Representations},
  year={2026}
}

@inproceedings{SAM2,
  title={{SAM 2}: Segment anything in images and videos},
  author={Ravi, Nikhila and Gabeur, Valentin and Hu, Yuan-Ting and Hu, Ronghang and Ryali, Chaitanya and Ma, Tengyu and Khedr, Haitham and R{\"a}dle, Roman and Rolland, Chloe and Gustafson, Laura and others},
  year={2025},
  booktitle={International Conference on Learning Representations}
}

@inproceedings{NVIDIA_Dynamic_Scene,
  title={Novel view synthesis of dynamic scenes with globally coherent depths from a monocular camera},
  author={Yoon, Jae Shin and Kim, Kihwan and Gallo, Orazio and Park, Hyun Soo and Kautz, Jan},
  booktitle={Proceedings of the IEEE/CVF Conference on Computer Vision and Pattern Recognition},
  pages={5336--5345},
  year={2020}
}

@article{Dycheck,
  title={Monocular dynamic view synthesis: A reality check},
  author={Gao, Hang and Li, Ruilong and Tulsiani, Shubham and Russell, Bryan and Kanazawa, Angjoo},
  journal={Advances in Neural Information Processing Systems},
  volume={35},
  pages={33768--33780},
  year={2022}
}

@inproceedings{DAVIS,
  title={A benchmark dataset and evaluation methodology for video object segmentation},
  author={Perazzi, Federico and Pont-Tuset, Jordi and McWilliams, Brian and Van Gool, Luc and Gross, Markus and Sorkine-Hornung, Alexander},
  booktitle={Proceedings of the IEEE Conference on Computer Vision and Pattern Recognition},
  pages={724--732},
  year={2016}
}

@article{HunyuanVideo,
  title={{HunyuanVideo}: A systematic framework for large video generative models},
  author={Kong, Weijie and Tian, Qi and Zhang, Zijian and Min, Rox and Dai, Zuozhuo and Zhou, Jin and Xiong, Jiangfeng and Li, Xin and Wu, Bo and Zhang, Jianwei and others},
  journal={arXiv preprint arXiv:2412.03603},
  year={2024}
}

@misc{Mochi,
      title={{Mochi 1}},
      author={{Genmo Team}},
      year={2024},
      publisher = {GitHub},
      journal = {GitHub repository},
      howpublished={\url{https://github.com/genmoai/models}}
}

@inproceedings{vbench,
  title={{VBench}: Comprehensive benchmark suite for video generative models},
  author={Huang, Ziqi and He, Yinan and Yu, Jiashuo and Zhang, Fan and Si, Chenyang and Jiang, Yuming and Zhang, Yuanhan and Wu, Tianxing and Jin, Qingyang and Chanpaisit, Nattapol and others},
  booktitle={Proceedings of the IEEE/CVF Conference on Computer Vision and Pattern Recognition},
  pages={21807--21818},
  year={2024}
}

@inproceedings{clipscore,
  title={{CLIPScore}: A reference-free evaluation metric for image captioning},
  author={Hessel, Jack and Holtzman, Ari and Forbes, Maxwell and Le Bras, Ronan and Choi, Yejin},
  booktitle={Proceedings of the 2021 Conference on Empirical Methods in Natural Language Processing},
  pages={7514--7528},
  year={2021}
}

@article{UnifiedReward,
  title={Unified reward model for multimodal understanding and generation},
  author={Wang, Yibin and Zang, Yuhang and Li, Hao and Jin, Cheng and Wang, Jiaqi},
  journal={arXiv preprint arXiv:2503.05236},
  year={2025}
}

@inproceedings{DiT,
  title={Scalable diffusion models with transformers},
  author={Peebles, William and Xie, Saining},
  booktitle={Proceedings of the IEEE/CVF International Conference on Computer Vision},
  pages={4195--4205},
  year={2023}
}

@inproceedings{LoRA,
    title={{LoRA}: Low-Rank Adaptation of Large Language Models},
    author={Edward J Hu and Yelong Shen and Phillip Wallis and Zeyuan Allen-Zhu and Yuanzhi Li and Shean Wang and Lu Wang and Weizhu Chen},
    booktitle={International Conference on Learning Representations},
    year={2022},
}

@article{Qwen2.5-VL,
  title={{Qwen2.5-VL} Technical Report},
  author={Bai, Shuai and Chen, Keqin and Liu, Xuejing and Wang, Jialin and Ge, Wenbin and Song, Sibo and Dang, Kai and Wang, Peng and Wang, Shijie and Tang, Jun and Zhong, Humen and Zhu, Yuanzhi and Yang, Mingkun and Li, Zhaohai and Wan, Jianqiang and Wang, Pengfei and Ding, Wei and Fu, Zheren and Xu, Yiheng and Ye, Jiabo and Zhang, Xi and Xie, Tianbao and Cheng, Zesen and Zhang, Hang and Yang, Zhibo and Xu, Haiyang and Lin, Junyang},
  journal={arXiv preprint arXiv:2502.13923},
  year={2025}
}

@article{Sa2VA,
  title={{Sa2VA}: Marrying {SAM2} with {MLLM} for Dense Grounded Understanding of Images and Videos},
  author={Yuan, Haobo and Li, Xiangtai and Zhang, Tao and Sun, Yueyi and Huang, Zilong and Xu, Shilin and Ji, Shunping and Tong, Yunhai and Qi, Lu and Feng, Jiashi and Yang, Ming-Hsuan},
  journal={IEEE Transactions on Pattern Analysis and Machine Intelligence},
  year={2026}
}

@article{UFM,
  title={{UFM}: A simple path towards unified dense correspondence with flow},
  author={Zhang, Yuchen and Keetha, Nikhil and Lyu, Chenwei and Jhamb, Bhuvan and Chen, Yutian and Qiu, Yuheng and Karhade, Jay and Jha, Shreyas and Hu, Yaoyu and Ramanan, Deva and others},
  journal={arXiv preprint arXiv:2506.09278},
  year={2025}
}

@article{gsplat,
  title={{gsplat}: An open-source library for {Gaussian} splatting},
  author={Ye, Vickie and Li, Ruilong and Kerr, Justin and Turkulainen, Matias and Yi, Brent and Pan, Zhuoyang and Seiskari, Otto and Ye, Jianbo and Hu, Jeffrey and Tancik, Matthew and Kanazawa, Angjoo},
  journal={Journal of Machine Learning Research},
  volume={26},
  number={34},
  pages={1--17},
  year={2025}
}

@inproceedings{LPIPS,
  title={Perceptual losses for real-time style transfer and super-resolution},
  author={Johnson, Justin and Alahi, Alexandre and Fei-Fei, Li},
  booktitle={European Conference on Computer Vision},
  pages={694--711},
  year={2016}
}

@article{vkitt2,
  title={Virtual {KITTI} 2},
  author={Cabon, Yohann and Murray, Naila and Humenberger, Martin},
  journal={arXiv preprint arXiv:2001.10773},
  year={2020}
}

@inproceedings{stereo4d,
  title={{Stereo4D}: Learning How Things Move in {3D} from Internet Stereo Videos}, 
  author={Jin, Linyi and Tucker, Richard and Li, Zhengqi and Fouhey, David and Snavely, Noah and Holynski, Aleksander},
  booktitle={Proceedings of the IEEE/CVF Conference on Computer Vision and Pattern Recognition},
  year={2025},
  pages={10497--10509}
}

@inproceedings{co3dv2,
  title={Common objects in {3D}: Large-scale learning and evaluation of real-life {3D} category reconstruction},
  author={Reizenstein, Jeremy and Shapovalov, Roman and Henzler, Philipp and Sbordone, Luca and Labatut, Patrick and Novotny, David},
  booktitle={Proceedings of the IEEE/CVF International Conference on Computer Vision},
  pages={10901--10911},
  year={2021}
}

@inproceedings{hypersim,
	title={{Hypersim}: A photorealistic synthetic dataset for holistic indoor scene understanding},
	author={Roberts, Mike and Ramapuram, Jason and Ranjan, Anurag and Kumar, Atulit and Bautista, Miguel Angel and Paczan, Nathan and Webb, Russ and Susskind, Joshua M},
	booktitle={Proceedings of the IEEE/CVF International Conference on Computer Vision},
	pages={10912--10922},
	year={2021}
}

@inproceedings{tartanair,
	title={{TartanAir}: A dataset to push the limits of visual {SLAM}},
	author={Wang, Wenshan and Zhu, Delong and Wang, Xiangwei and Hu, Yaoyu and Qiu, Yuheng and Wang, Chen and Hu, Yafei and Kapoor, Ashish and Scherer, Sebastian},
	booktitle={IEEE/RSJ International Conference on Intelligent Robots and Systems},
	pages={4909--4916},
	year={2020},
}

@inproceedings{pointodyssey,
	title={{PointOdyssey}: A large-scale synthetic dataset for long-term point tracking},
	author={Zheng, Yang and Harley, Adam W and Shen, Bokui and Wetzstein, Gordon and Guibas, Leonidas J},
	booktitle={Proceedings of the IEEE/CVF International Conference on Computer Vision},
	pages={19855--19865},
	year={2023}
}

@inproceedings{bedlam,
  title = {{BEDLAM}: A Synthetic Dataset of Bodies Exhibiting Detailed Lifelike Animated Motion},
  author = {Black, Michael J. and Patel, Priyanka and Tesch, Joachim and Yang, Jinlong}, 
  booktitle={Proceedings of the IEEE/CVF Conference on Computer Vision and Pattern Recognition},
  pages = {8726--8737},
  year = {2023}
}

@inproceedings{dynamicreplica,
  title={{DynamicStereo}: Consistent Dynamic Depth from Stereo Videos},
  author={Nikita Karaev and Ignacio Rocco and Benjamin Graham and Natalia Neverova and Andrea Vedaldi and Christian Rupprecht},
  booktitle={Proceedings of the IEEE/CVF Conference on Computer Vision and Pattern Recognition},
  year={2023},
  pages={13229--13239}
}

@inproceedings{mvssynth,
	title={{DeepMVS}: Learning multi-view stereopsis},
	author={Huang, Po-Han and Matzen, Kevin and Kopf, Johannes and Ahuja, Narendra and Huang, Jia-Bin},
	booktitle={Proceedings of the IEEE Conference on Computer Vision and Pattern Recognition},
	pages={2821--2830},
	year={2018}
}

@article{omniworld,
      title={{OmniWorld}: A Multi-Domain and Multi-Modal Dataset for {4D} World Modeling}, 
      author={Yang Zhou and Yifan Wang and Jianjun Zhou and Wenzheng Chang and Haoyu Guo and Zizun Li and Kaijing Ma and Xinyue Li and Yating Wang and Haoyi Zhu and Mingyu Liu and Dingning Liu and Jiange Yang and Zhoujie Fu and Junyi Chen and Chunhua Shen and Jiangmiao Pang and Kaipeng Zhang and Tong He},
      journal={arXiv preprint arXiv:2509.12201},
      year={2025}
}

@inproceedings{ase,
  author    = {Pan, Xiaqing and Charron, Nicholas and Yang, Yongqian and Peters, Scott and Whelan, Thomas and Kong, Chen and Parkhi, Omkar and Newcombe, Richard and Ren, Yuheng (Carl)},
  title     = {Aria {D}igital {T}win: A New Benchmark Dataset for Egocentric 3{D} Machine Perception},
  booktitle = {Proceedings of the IEEE/CVF International Conference on Computer Vision},
  year      = {2023},
  pages     = {20133-20143}
}

@InProceedings{spring,
    author    = {Lukas Mehl and Jenny Schmalfuss and Azin Jahedi and Yaroslava Nalivayko and Andr\'es Bruhn},
    title     = {{Spring}: A High-Resolution High-Detail Dataset and Benchmark for Scene Flow, Optical Flow and Stereo},
    booktitle = {Proceedings of the IEEE/CVF Conference on Computer Vision and Pattern Recognition},
    year      = {2023},
  pages={4981--4991}
}

@inproceedings{waymo, 
	title = {Scalability in Perception for Autonomous Driving: {Waymo} Open Dataset}, 
	author = {Sun, Pei and Kretzschmar, Henrik and Dotiwalla, Xerxes and Chouard, Aurelien and Patnaik, Vijaysai and Tsui, Paul and Guo, James and Zhou, Yin and Chai, Yuning and Caine, Benjamin and Vasudevan, Vijay and Han, Wei and Ngiam, Jiquan and Zhao, Hang and Timofeev, Aleksei and Ettinger, Scott and Krivokon, Maxim and Gao, Amy and Joshi, Aditya and Zhang, Yu and Shlens, Jonathon and Chen, Zhifeng and Anguelov, Dragomir}, 
	booktitle = {Proceedings of the IEEE/CVF Conference on Computer Vision and Pattern Recognition}, 
	year = {2020},
  pages={2446--2454}
}

@article{arkitscenes,
  title={{ARKitScenes}: A diverse real-world dataset for {3D} indoor scene understanding using mobile {RGB-D} data},
  author={Baruch, Gilad and Chen, Zhuoyuan and Dehghan, Afshin and Dimry, Tal and Feigin, Yuri and Fu, Peter and Gebauer, Thomas and Joffe, Brandon and Kurz, Daniel and Schwartz, Arik and others},
  journal={arXiv preprint arXiv:2111.08897},
  year={2021}
}

@inproceedings{scannetpp,
  title={{ScanNet++}: A High-Fidelity Dataset of {3D} Indoor Scenes},
  author={Yeshwanth, Chandan and Liu, Yueh-Cheng and Nie{\ss}ner, Matthias and Dai, Angela},
  booktitle = {Proceedings of the IEEE/CVF International Conference on Computer Vision},
  year={2023},
  pages={12--22}
}

@inproceedings{dl3dv,
  title={{DL3DV-10K}: A large-scale scene dataset for deep learning-based {3D} vision},
  author={Ling, Lu and Sheng, Yichen and Tu, Zhi and Zhao, Wentian and Xin, Cheng and Wan, Kun and Yu, Lantao and Guo, Qianyu and Yu, Zixun and Lu, Yawen and others},
  booktitle={Proceedings of the IEEE/CVF Conference on Computer Vision and Pattern Recognition},
  pages={22160--22169},
  year={2024}
}

@inproceedings{sintel,
  title={A naturalistic open source movie for optical flow evaluation},
  author={Butler, Daniel J and Wulff, Jonas and Stanley, Garrett B and Black, Michael J},
  booktitle={European Conference on Computer Vision},
  pages={611--625},
  year={2012}
}

@inproceedings{bonn,
  title={{ReFusion}: {3D} reconstruction in dynamic environments for {RGB-D} cameras exploiting residuals},
  author={Palazzolo, Emanuele and Behley, Jens and Lottes, Philipp and Gigu{\`e}re, Philippe and Stachniss, Cyrill},
  booktitle={IEEE/RSJ International Conference on Intelligent Robots and Systems},
  pages={7855--7862},
  year={2019}
}

@article{kitti,
  title={Vision meets robotics: The {KITTI} dataset},
  author={Geiger, Andreas and Lenz, Philip and Stiller, Christoph and Urtasun, Raquel},
  journal={The International Journal of Robotics Research},
  volume={32},
  number={11},
  pages={1231--1237},
  year={2013},
}

@article{ViGeo,
  title={Towards Consistent Video Geometry Estimation},
  author={Yu, Zhu and Gao, Jingnan and Zhang, Runmin and Qiu, Lingteng and Zhao, Zhengyi and Peng, Rui and Yan, Yichao and Qiu, Kejie and Zhu, Siyu and Dong, Zilong and Cao, Si-Yuan and Shen, Hui-Liang},
  journal={arXiv preprint arXiv:2605.30060},
  year={2026}
}

@article{VGGT-omega,
  title = {{VGGT}-{$\Omega$}},
  author={Wang, Jianyuan and Chen, Minghao and Zhang, Shangzhan and Karaev, Nikita and Sch{\"o}nberger, Johannes and Labatut, Patrick and Bojanowski, Piotr and Novotny, David and Vedaldi, Andrea and Rupprecht, Christian},
  journal={arXiv preprint arXiv:2605.15195},
  year={2026}
}

@article{GAE,
  title={{GAE}: Learning a geometry-native latent space for {3D}-consistent world generation},
  author={Lu, Jiahao and Yin, Minghao and Hu, Wenbo and Liu, Hengyu and Zhao, Wang and Yeung, Sai-Kit and Shan, Ying and Liu, Yuan},
  journal={arXiv preprint arXiv:2609.24981},
  year={2026}
}

@article{PixWorld,
  title={{PixWorld}: Unifying {3D} scene generation and reconstruction in pixel space},
  author={Gao, Sensen and Wang, Zhaoqing and Cao, Qihang and Yu, Dongdong and Wang, Changhu and Bian, Jia-Wang},
  journal={arXiv preprint arXiv:2607.05373},
  year={2026}
}

@inproceedings{VGGRPO,
  title={{VGGRPO}: Towards world-consistent video generation with {4D} latent reward},
  author={An, Zhaochong and Kupyn, Orest and Uscidda, Th{\'e}o and Colaco, Andrea and Ahuja, Karan and Belongie, Serge and Gonzalez-Franco, Mar and Tintore Gazulla, Marta},
  booktitle={European Conference on Computer Vision},
  pages={305--322},
  year={2026},
  organization={Springer}
}

@article{BeyondPixels,
  title={{Beyond Pixels}: From video priors to {4D} worlds},
  author={Liu, Zihao and Shen, Xiaolong and Zhou, Zhenglin and Quan, Ruijie and Yang, Yi},
  journal={arXiv preprint arXiv:2608.10744},
  year={2026}
}

@inproceedings{AEON,
  title={{AEON}: Unifying video and {3D} world models},
  author={Zhang, Weiqi and Zhang, Wenyuan and Liu, Yu-Shen and Zhou, Junsheng},
  year={2026},
  booktitle={Advances in Neural Information Processing Systems}
}

@inproceedings{C3G,
  title={Learning compact {3D} representations from feed-forward novel view synthesis},
  author={An, Honggyu and Jung, Jaewoo and Kim, Mungyeom and Kim, Chaehyun and Jeon, Minkyeong and Han, Jisang and Fukuda, Kazumi and Narihira, Takuya and Ko, Hyunah and Kim, Junsu and Hong, Sunghwan and Mitsufuji, Yuki and Kim, Seungryong},
  year={2026},
  booktitle={Proceedings of the IEEE/CVF Conference on Computer Vision and Pattern Recognition},
  pages={363--373}
}

@article{C4G,
  title={Learning global motion with compact {Gaussians} for feed-forward {4D} reconstruction},
  author={Kim, Mungyeom and Jeon, Minkyeong and An, Honggyu and Jung, Jaewoo and Ko, Hyuna and Han, Jisang and Yu, Hyeonseo and Shin, Donghwan and Hong, Sunghwan and Narihira, Takuya and Fukuda, Kazumi and Mitsufuji, Yuki and Kim, Seungryong},
  journal={arXiv preprint arXiv:2605.31595},
  year={2026}
}

@article{GSExplorer,
  title={{GSExplorer}: Interactive world generation with a persistent {3D} {Gaussian} world state},
  author={Yu, Sicheng and Liu, Liu and Wang, Xinjie and Wang, Xinjiang and Ding, Taojun and Hu, Wenhao and Zhao, Beizhen and Yan, Puxin and Su, Zhizhong and Wang, Hao},
  journal={arXiv preprint arXiv:2611.01543},
  year={2026}
}

@article{IGGT4D,
  title={{IGGT4D}: Streaming {4D} instance-grounded geometry transformer},
  author={Zou, Zhengyu and Li, Hao and Jiao, Kuixuan and Liu, Liu and Xiao, Tingyang and Zhou, Xiaolin and Hong, Fangzhou and Su, Zhizhong and Zhang, Dingwen and Liu, Ziwei},
  journal={arXiv preprint arXiv:2607.19228},
  year={2026}
}

@inproceedings{Uni3R,
  title={{Uni3R}: Unified {3D} reconstruction and semantic understanding via generalizable {Gaussian} splatting from unposed multi-view images},
  author={Sun, Xiangyu and Jiang, Haoyi and Liu, Liu and Nam, Seungtae and Kang, Gyeongjin and Wang, Xinjie and Sui, Wei and Su, Zhizhong and Liu, Wenyu and Wang, Xinggang and Park, Eunbyung},
  booktitle={Proceedings of the IEEE/CVF Conference on Computer Vision and Pattern Recognition},
  pages={33280--33290},
  year={2026}
}

@article{PreF3R,
  title={{PreF3R}: Pose-free feed-forward {3D} {Gaussian} splatting from variable-length image sequence},
  author={Chen, Zequn and Yang, Jiezhi and Yang, Heng},
  journal={arXiv preprint arXiv:2411.16877},
  year={2024}
}

@inproceedings{MORE_4D,
  title={{MoRe}: Motion-aware feed-forward {4D} reconstruction transformer},
  author={Fang, Juntong and Chen, Zequn and Zhang, Weiqi and Di, Donglin and Zhang, Xuancheng and Yang, Chengmin and Liu, Yu-Shen},
  year={2026},
  booktitle={Proceedings of the IEEE/CVF Conference on Computer Vision and Pattern Recognition},
  pages={28914--28924}
}

@inproceedings{Anchor3DGS,
  title = {{Anchor3DGS}: Feed-forward {3D} {Gaussian} splatting with compact anchor-based representation},
  author = {Ye, Sheng and Lin, Jiangke and Wang, Fudong and Yuan, Yi and Dong, Zhen-Hui and Liu, Yong-Jin},
  year = {2026},
  booktitle = {Advances in Neural Information Processing Systems}
}
}

\end{document}